\documentclass[twocolumn,final]{svjour3}

\usepackage[hyphens]{url}
\usepackage[british]{babel}
\usepackage{graphicx}
\usepackage{natbib}
\usepackage{amssymb}
\usepackage{amsmath}

\newcommand{\mref}[1]{(\ref{#1})}
\newcommand{\given}{\mid}
\newcommand{\argmin}{\operatornamewithlimits{argmin}}
\newcommand{\OLS}[1]{\widehat{#1}_{\mathrm{OLS}}}
\newcommand{\NCLM}[1]{\widehat{#1}_{\mathrm{NCLM}}}
\newcommand{\FRRM}[1]{\widehat{#1}_{\mathrm{FRRM}}}
\newcommand{\FGRRM}[1]{\widehat{#1}_{\mathrm{FGRRM}}}
\newcommand{\y}{\mathbf{y}}
\newcommand{\wy}{\widehat{\y}}
\newcommand{\Xm}{\mathbf{X}}
\newcommand{\Sm}{\mathbf{S}}
\newcommand{\B}{\mathbf{B}}
\newcommand{\BOLS}{\OLS{\B}}
\newcommand{\BRD}{\widehat{\B}_{\lambda}}
\newcommand{\T}{\mathrm{T}}
\newcommand{\U}{\mathbf{U}}
\newcommand{\wU}{\widehat{\U}}
\newcommand{\Zm}{\mathbf{0}}
\newcommand{\I}{\mathbf{I}}
\newcommand{\balpha}{\boldsymbol{\alpha}}
\newcommand{\bbeta}{\boldsymbol{\beta}}
\newcommand{\be}{\boldsymbol{\varepsilon}}
\newcommand{\rs}{R^2_\Sm}
\newcommand{\E}{\operatorname{E}}
\newcommand{\VAR}{\operatorname{VAR}}
\newcommand{\COV}{\operatorname{COV}}
\newcommand{\COR}{\operatorname{COR}}
\newcommand{\aVa}{\balpha^\T \VAR(\Sm) \balpha}
\newcommand{\bVb}{\bbeta^\T \VAR(\wU) \bbeta}
\newcommand{\st}{\operatorname{s.t.}}
\newcommand{\diag}{\operatorname{diag}}
\newcommand{\A}{\mathbf{A}}
\newcommand{\tU}{\widetilde{\mathbf{U}}}
\newcommand{\tu}{\widetilde{U}_i}
\newcommand{\wu}{\widehat{U}_i}
\newcommand{\bmu}{\boldsymbol{\mu}}
\newcommand{\eeta}{\boldsymbol{\eta}}
\newcommand{\weta}{\widehat{\eeta}}
\newcommand{\pphi}{\boldsymbol{\phi}}
\newcommand{\reo}{R^2_\mathrm{EO}}
\newcommand{\lr}{\lambda(r)}
\newcommand{\II}{\mathbf{I}}
\newcommand{\LLambda}{\boldsymbol{\mathrm{\Lambda}}}
\newcommand{\zcov}{|\COV(\weta, S_i)|}
\newcommand{\zcovy}{|\COV(\wy, S_i)|}
\newcommand{\ui}{\mathbf{u}_i}
\newcommand{\uj}{\mathbf{u}_j}
\newcommand{\si}{\mathbf{s}_i}
\newcommand{\sj}{\mathbf{s}_j}
\newcommand{\DIF}{D_{\mathrm{IF}}}

\newenvironment{continued}{%
  \addtocounter{example}{-1}
  \begin{example}[continued]
}{%
  \end{example}
}

\title{Achieving Fairness with a Simple Ridge Penalty}
\subtitle{}

\author{Marco Scutari \and Francesca Panero \and Manuel Proissl}

\institute{
  M. Scutari \at
    Istituto Dalle Molle di Studi sull'Intelligenza Artificiale \mbox{(IDSIA)},
    Lugano, Switzerland \\
    \email{scutari@idsia.ch}
  \and F. Panero \at
    London School of Economics; formerly
    Department of Statistics, University of Oxford, Oxford, United Kingdom \\
    \email{f.panero@lse.ac.uk}
  \and M. Proissl \at
    UBS, Data \& Analytics Center of Excellence, Zurich, Switzerland \\
    \email{manuel.proissl@ubs.com}
}

\begin{document}
\maketitle

% do not put mathematical notation or abbreviations.
\begin{abstract}
  In this paper we present a general framework for estimating regression models
  subject to a user-defined level of fairness. We enforce fairness as a model
  selection step in which we choose the value of a ridge penalty to control the
  effect of sensitive attributes. We then estimate the parameters of the model
  conditional on the chosen penalty value. Our proposal is mathematically
  simple, with a solution that is partly in closed form, and produces estimates
  of the regression coefficients that are intuitive to interpret as a function
  of the level of fairness. Furthermore, it is easily extended to generalised
  linear models, kernelised regression models and other penalties; and it can
  accommodate multiple definitions of fairness.

  We compare our approach with the regression model from \citet{komiyama}, which
  implements a provably-optimal linear regression model; and with the fair
  models from \citet{zafar}. We evaluate these approaches empirically on six
  different data sets, and we find that our proposal provides better goodness of
  fit and better predictive accuracy for the same level of fairness. In
  addition, we highlight a source of bias in the original experimental
  evaluation in \citet{komiyama}.
\end{abstract}

\section{Introduction}
\label{sec:intro}

Machine learning models are increasingly being used in applications where it is
crucial to ensure the accountability and fairness of the decisions made on the
basis of their outputs: some examples are criminal justice \citep{heidari},
credit risk modelling \citep{credit} and screening job applications
\citep{jobs}. In such cases, we are required to ensure that we are not
discriminating individuals based on sensitive attributes such as gender and
race, leading to disparate treatment of specific groups. At the same time, we
would like to achieve the best possible predictive accuracy from other
predictors.

The task of defining a non-discriminating treatment, though, does not come
without challenges. The concept of fairness itself, in fact, has been
characterised in different ways depending on the context. From an ethical and
legal perspective, for example, it might depend on the type of distortion we
wish to limit, which in turns varies with the type of application. Sometimes, we
want to limit the adverse bias against a specific group, while in other
instances we wish to protect single individuals. Alongside the legal and
philosophical research debate, institutional regulations on the use of
algorithms in society have been proposed in the last decade: for a comparison
among the US, EU and UK regulations see \citet{cath}. The European Commission
has recently released the first legal framework for the use of artificial
intelligence \citep{eu2021}, which is now under revision by the member states.

At the same time, there has been a growing interest towards fairness-preserving
methods in the machine learning literature. From a statistical perspective,
different characterisations of fairness translate into different probabilistic
models, which must then take into account the characteristics of the data they
may be applied to (say, whether the variables are continuous or categorical, or
whether a reliable ground truth is available or not). Despite this variety, all
characterisations of fairness follow one of the two following approaches:
\textit{individual fairness} or \textit{group fairness} \citep{delbarrio}. The
former requires that individuals that are similar receive similar predictions,
while the latter aims at obtaining predictions that are similar across the
groups identified by the sensitive variables.

Group fairness has been explored the most in the literature. When defined as
\textit{statistical} or \textit{demographic parity}, it requires that
predictions and the sensitive variables are independent. If $\Xm$ is a matrix of
predictors, $\Sm$ is a matrix of sensitive attributes, $\y$ is the response
variable and $\wy$ are the predictions provided by some model, statistical parity
translates into $\wy$ being independent of $\Sm$. Usually the requirement of
complete independence is too strong for practical applications, and it is
relaxed into a constraint that limits the strength of the dependence between
$\Sm$ and $\wy$. Statistical parity is a good definition when a reliable ground
truth is not available: otherwise, a perfect classifier on a data set which is
unbalanced in the outcome across groups would possibly not satisfy the
definition. While this is usually seen as a weakness of this fairness
definition, we must remember that historical data often display some sort of
bias and allowing the perfect classifier to score optimally in the chosen
metric would mean to preserve it in future decisions. In light of this,
definitions that do not rely on ground truth are known as ``bias-transforming'',
while notions that condition on the truth are known as ``bias-preserving''
\citep{wachter}. A common bias-preserving definition of fairness is
\textit{equality of opportunity}, which requires the predictions $\wy$ to be
independent from the sensitive attributes $\Sm$ after conditioning on the ground
truth $\y$. In the case of a binary classifier and a single categorical
sensitive variable, it is commonly known as \textit{equality of odds} and
translates into having the same false positive and negative rates across
different groups.

Learning fair black-box machine learning models such as deep neural networks
provides many hard challenges \citep[see, for instance,][]{choras} that are
currently being investigated. For this reason, a large part of the literature
focuses on simpler models. In many settings, such models are preferable because
there are not enough data to train a deep neural network, because of
computational limitations, or because they are more interpretable. Most such
research focuses on classification models. For instance, \citet{woodworth}
investigated equality of odds for a binary sensitive attribute; \cite{zafar}
investigated the unfairness of the decision boundary in logistic regression and
support-vector machines under statistical parity; \citet{silva-nips} explored
counterfactual fairness in graphical models; \citet{reductions} used
ensembles of logistic regressions and gradient-boosted trees to reduce fair
classification to a sequence of cost-sensitive classification problems. For a
review on the vast field of fairness notions and methods, see \citet{mehrabi},
\citet{delbarrio} and \citet{pessach}. In our work, we will focus on models that
satisfy statistical parity.

Fair regression models (with a continuous response variable) have not been
investigated in the literature as thoroughly as classifiers. \citet{fukuchi}
considered a generative model that is neutral to a finite set of viewpoints.
\citet{calders} focused on discrete sensitive attributes that may be used to
cluster observations. \citet{suay} used kernels as a regulariser to enforce
fairness while allowing non-linear associations and dimensionality reduction.
\citet{grouploss} chose to bound the regression error within an allowable limit
for each group defined by the sensitive attributes; \citet{berk} achieved a
similar effect using individual and group penalty terms. \cite{chzhen} used
model recalibration on a discretised transform of the response to leverage
fairness characterisations used in classification models with a single binary
sensitive attribute. \citet{mary} used the notion of R\'enyi correlation to
propose two methods that achieve statistical parity and equality of odds.
\citet{steinberg} implemented a similar idea with mutual information.

\citet{komiyama} proposed a quadratic optimisation approach for fair linear
regression models that constrains least squares estimation by bounding the
relative proportion of the variance explained by the sensitive attributes,
falling into the statistical parity framework. In contrast with the approaches
mentioned above, both predictors and sensitive attributes are allowed to be
continuous as well as discrete; any number of predictors and sensitive
attributes can be included in the model; and the level of fairness can be
controlled directly by the user, without the need of model calibration to
estimate it empirically. In the following we call this approach NCLM (as in
\textit{non-convex linear model}). This approach comes with theoretical
optimality guarantees. However, it produces regression coefficient estimates
that are not in closed form and whose behaviour is not easy to interpret with
respect to the level of fairness; and it is difficult to extend it beyond linear
regression models.

These limitations motivated us to propose a simpler \textit{fair ridge
regression model} (FRRM) which is easier to estimate, to interpret and to
extend. At the same time, we wanted to match the key strengths that distinguish
\cite{komiyama} from earlier work:
\begin{enumerate}
  \item the ability to model any combination of discrete and continuous
    predictors as well as sensitive attributes;
  \item the ability to control fairness directly via a tuning parameter with an
    intuitive, real-world interpretation.
\end{enumerate}
We achieve these aims by separating model selection and model estimation.
Firstly, we choose the ridge penalty to achieve the desired level of fairness.
Secondly, we estimate the model parameters conditional on the chosen penalty
value. This is in contrast with other methods in the literature that do not have
a separate model selection phase.

The paper is laid out as follows. In Section \ref{sec:nclm} we briefly review
the NCLM approach from \citet{komiyama}, its formulation as an optimisation
problem (Section \ref{sec:ncopt}), and we highlight a source of bias in its
original experimental validation (Section \ref{sec:bias}). In Section
\ref{sec:frrm} we discuss our proposal, FRRM, including its practical
implementation (Section \ref{sec:fropt}). We also discuss an analytical,
closed-form estimate for ridge penalty and for the regression coefficients of
the sensitive attributes in Section \ref{sec:analytical}. We discuss several
possible extensions of FRRM, including that to generalised linear models (FGRRM)
and to different definitions of fairness in Section \ref{sec:extensions}. In
Section \ref{sec:simulations} we compare FRRM with NCLM and with the approach
proposed by \citet{zafar}, investigating both linear (Section
\ref{sec:linearsim}) and logistic (Section \ref{sec:logisticsim}) regression
models. We also consider the models from \citet{steinberg} and
\citet{reductions} in Section \ref{sec:othersim}, insofar as they can be
compared to FRRM given their limitations. Finally, we discuss the results in
Section \ref{sec:conclusions}.

\section{A Nonconvex Optimisation Approach to Fairness}
\label{sec:nclm}

Let $\Xm$ be a matrix of predictors, $\Sm$ be a matrix of sensitive attributes
and $\y$ be a continuous response variable. Without loss of generality, we
assume that all variables in $\Xm$, $\Sm$ and $\y$ are centred, and that any
categorical variables in $\Xm$ and $\Sm$ have been replaced with their one-hot
encoding. \citet{komiyama} start by removing the association between $\Xm$ and
$\Sm$ using the auxiliary multivariate linear regression model
\begin{equation*}
  \Xm = \B^\T \Sm + \U.
\end{equation*}
They estimate the regression coefficients $\B$ by ordinary least squares (OLS)
as $\BOLS = (\Sm^\T \Sm)^{-1} \Sm^\T \Xm$, thus obtaining the residuals
\begin{equation}
  \wU = \Xm - \BOLS^\T \Sm.
\label{eq:prep}
\end{equation}
Due to the properties of OLS, $\Sm$ and $\wU$ are orthogonal and
$\COV(\Sm, \wU) = \Zm$, where $\Zm$ is a matrix of zeroes. $\BOLS^\T \Xm$
can then be interpreted as the component of $\Xm$ that is explained by $\Sm$,
and $\wU$ as the component of $\Xm$ that cannot be explained by $\Sm$ (the
de-correlated predictors).

\citet{komiyama} then define their main predictive model as the OLS regression
\begin{equation}
  \y = \Sm \balpha + \wU \bbeta + \be,
\label{eq:komiyama-main}
\end{equation}
where $\balpha$ is the vector of the coefficients associated with the sensitive
attributes $\Sm$, and $\bbeta$ is associated with the de-correlated predictors
$\wU$. In keeping with classical statistics, they measure the goodness-of-fit of
the model with the coefficient of determination $R^2(\balpha, \bbeta)$, which
can be interpreted as the proportion of variance explained by the model.
Furthermore, $R^2(\balpha, \bbeta)$ is also proportional to the cross-entropy of
the model, which is a function of $1 - R^2(\balpha, \bbeta)$. Since $\Sm$ and
$\wU$ are orthogonal, and since \mref{eq:komiyama-main} is fitted with OLS,
$R^2(\balpha, \bbeta)$ decomposes as
\begin{multline}
  R^2(\balpha, \bbeta) = \frac{\VAR(\wy)}{\VAR(\y)}
      = \frac{\VAR(\Sm \balpha + \wU \bbeta)}
             {\VAR(\Sm \balpha + \wU \bbeta + \be)} = \\
        \frac{\aVa + \bVb}{\aVa + \bVb + \VAR(\be)},
\label{eq:r2}
\end{multline}
where $\wy$ are the fitted values produced by OLS, and $\VAR(\wU)$, $\VAR(\Sm)$
are the covariance matrices of $\Xm$ and $\Sm$, respectively. Both matrices are
assumed to be full-rank. The proportion of the overall explained variance that
is attributable to the sensitive attributes then is
\begin{multline}
  \rs(\balpha, \bbeta) = \frac{\VAR(\Sm\balpha)}{\VAR(\wy)} = \\
    \frac{\aVa}{\aVa + \bVb}.
\label{eq:bound}
\end{multline}
\citet{komiyama} choose to bound $\rs(\balpha, \bbeta)$ to a value $r \in [0,
1]$ that determines how fair the model is. Setting $r = 0$ corresponds to a
completely fair model (that is, statistical parity: $\wy$ is independent from
$\Sm$) because it implies $\aVa = 0$, which can only be true if all regression
coefficients $\balpha$ are equal to zero. On the other hand, setting $r = 1$
means the constraint is always satisfied because $\rs(\balpha, \bbeta) \leqslant
1$ by construction.

\subsection{The Optimisation Problem}
\label{sec:ncopt}

Fitting \mref{eq:komiyama-main} subject to $\rs(\balpha, \bbeta) \leqslant r$
by OLS can be formally be written as
\begin{equation}
  \begin{aligned}
    &\min_{\balpha, \bbeta} & &\E\left((\y - \wy)^2\right) \\
    &\st  & &\rs(\balpha, \bbeta) \leqslant r
  \end{aligned}
\label{eq:general}
\end{equation}
which in light of \mref{eq:r2} and \mref{eq:bound} takes the form
\begin{equation}
  \begin{aligned}
    &\min_{\balpha, \bbeta} &
          &\aVa + \bVb - \\
    &     & & \qquad\qquad 2\left( \E(\y \Sm^\T \balpha) + \E(\y \wU^\T \bbeta) \right)\\
    &\st  & &(1 - r)\aVa - r \bVb \leqslant 0.
  \end{aligned}
\label{eq:qcqp}
\end{equation}
The optimisation in \mref{eq:qcqp} is a quadratic programming problem subject to
quadratic constraints, and it is not convex in $(\balpha, \bbeta)$. Instead of
solving \mref{eq:qcqp}, \citet{komiyama} converts it into the convex quadratic
problem
\begin{equation}
  \begin{aligned}
    &\min_{\balpha, \bbeta, \gamma}& &\gamma \\
    &\st &
      &\begin{bmatrix} \balpha^\T & \bbeta^\T \end{bmatrix}
       \begin{bmatrix}
         \VAR(\Sm)  & \Zm \\
         \Zm        & \VAR(\wU)
       \end{bmatrix}
       \begin{bmatrix} \balpha \\ \bbeta \end{bmatrix} - \\
    &   &
      &\qquad\qquad 2 \begin{bmatrix}
         \E(\y\Sm) & \E(\y\wU)
       \end{bmatrix}
       \begin{bmatrix} \balpha \\ \bbeta \end{bmatrix} -
       \gamma \leqslant 0, \\
    &   &
      &\begin{bmatrix} \balpha^\T & \bbeta^\T \end{bmatrix}
       \begin{bmatrix}
         \VAR(\Sm)/r  & \Zm \\
         \Zm          & \Zm
       \end{bmatrix}
       \begin{bmatrix} \balpha \\ \bbeta \end{bmatrix} - \\
    &   &
      &\qquad\qquad 2 \begin{bmatrix}
         \E(\y\Sm) & \E(\y\wU)
       \end{bmatrix}
       \begin{bmatrix} \balpha \\ \bbeta \end{bmatrix} -
       \gamma \leqslant 0,
  \end{aligned}
\label{eq:gamma}
\end{equation}
using previous results from \cite{yamada}. ($\gamma$ is an auxiliary parameter
without a real-world interpretation.) The relaxed problem in \mref{eq:gamma}
yields an optimal solution $(\NCLM{\balpha}, \NCLM{\bbeta})$ for \mref{eq:qcqp},
under the assumptions discussed earlier as well as those in \cite{yamada}. It
also has two additional favourable properties: it can be solved by off-the-shelf
optimisers (the authors used Gurobi) and it can be extended by replacing
$\VAR(\Sm)$ and $\VAR(\wU)$ with more complex estimators than the respective
empirical covariance matrices. Two examples covered in \citet{komiyama} are the
use of kernel transforms to capture non-linear relationship and regularisation
adding a ridge penalty term.

\subsection{Avoiding Bias in the Auxiliary Model}
\label{sec:bias}

A key assumption that underlies the results in the previous section is the use
of OLS in creating the de-correlated predictors in \mref{eq:prep}: it ensures
that $\wU$ is orthogonal to $\Sm$ and therefore that is does not contain any
information from the sensitive attributes. However, \citet{komiyama} in their
experimental section state that ``The features $\wU$ were built from $\Xm$ by
de-correlating it from $\Sm$ by using regularised least squares
regression''\footnote{We substituted their notation with ours for clarity.},
where the ``regularised least squares regression'' is a ridge regression model.
This divergence from the theoretical construction leading to \mref{eq:gamma}
introduces bias in the model by making $\wU$ correlated to $\Sm$ in proportion
to the amount of regularisation.

As noted in \citet{ridgenotes}, the residuals in a ridge regression are not
orthogonal to the fitted values for any penalisation coefficient $\lambda > 0$.
Let $\tU$ be the ridge estimate of $\U$, that is, $\tU = \Xm - \BRD \Xm$. Let
$X_i$ be the $i$th column of $\Xm$ (that is, one of the predictors) and $\tu$ be
the corresponding column of $\tU$. Then
\begin{equation*}
  \tu = X_i - \Sm (\Sm^\T \Sm + \lambda \I)^{-1} \Sm^\T X_i
\end{equation*}
while the corresponding OLS estimate from $\wU$ is
\begin{equation*}
  \wu = X_i - \Sm (\Sm^\T \Sm)^{-1} \Sm^\T X_i.
\end{equation*}
Their difference is
\begin{align}
  \tu - \wu
    &= \Sm (\Sm^\T \Sm)^{-1} \Sm^\T X_i
         - \Sm (\Sm^\T \Sm + \lambda \I)^{-1} \Sm^\T X_i \notag \\
    &= \Sm \left[ (\Sm^\T \Sm)^{-1} -
         (\Sm^\T \Sm + \lambda \I)^{-1}\right] \Sm^\T X_i.
\label{eq:udiff}
\end{align}
Given the spectral decomposition $\Sm^\T \Sm = \A \LLambda \A^\T$, where
\mbox{$\LLambda = \diag(l_j)$}, \mref{eq:udiff} can be rewritten as
\begin{align*}
  \tu - \wu
    &= \Sm \left[ \A \LLambda^{-1} \A^\T -
         (\A \LLambda \A^\T + \lambda\I)^{-1}\right] \Sm^\T X_i \\
    &= \Sm \left[ \A
         \diag\left( \frac{1}{l_j} - \frac{1}{l_j + \lambda}\right) \A^\T
       \right] \Sm^\T X_i,
\end{align*}
thus giving
\begin{align*}
  \Sm^\T(\tu - \wu)
    &= \Sm^\T \Sm \left[ \A
         \diag\left( \frac{1}{l_j} - \frac{1}{l_j + \lambda}\right) \A^\T
       \right] \Sm^\T X_i \\
    &= \A \LLambda \A^\T \A
         \diag\left( \frac{1}{l_j} - \frac{1}{l_j + \lambda}\right) \A^\T
       \Sm^\T X_i \\
    &= \A \diag\left( 1 - \frac{l_j}{l_j + \lambda}\right) \A^\T
       \Sm^\T X_i.
\end{align*}
Since $\Sm^\T \tu = \Sm^\T (\tu - \wu)$ due to $\Sm^\T \wu = \Zm$, and
\mbox{$\COV(\Sm, \tu) \propto \Sm^\T \wu$}, we have that $\COV(\Sm, \tu)$
vanishes as $\lambda \to 0$ and $\BRD \to \BOLS$. On the other hand,
$|\COV(\Sm, \tu)|$ becomes increasingly large as $\lambda \to \infty$,
eventually reaching $\COV(\Sm, X_i)$. If we replace $\wU$ with $\tU$, the
denominator of $\rs(\balpha, \bbeta)$ then becomes
\begin{align*}
  \VAR(\widetilde{\y})
    &= \VAR(\Sm\balpha + \tU\bbeta) \\
    &= \VAR(\Sm\balpha) + \VAR(\tU\bbeta) - 2\COV(\Sm\balpha, \tU\bbeta)
\end{align*}
which can be either larger or smaller than that of the $\rs(\balpha, \bbeta)$ in
\mref{eq:bound}.

\begin{figure}[t]
  \includegraphics[width=\linewidth]{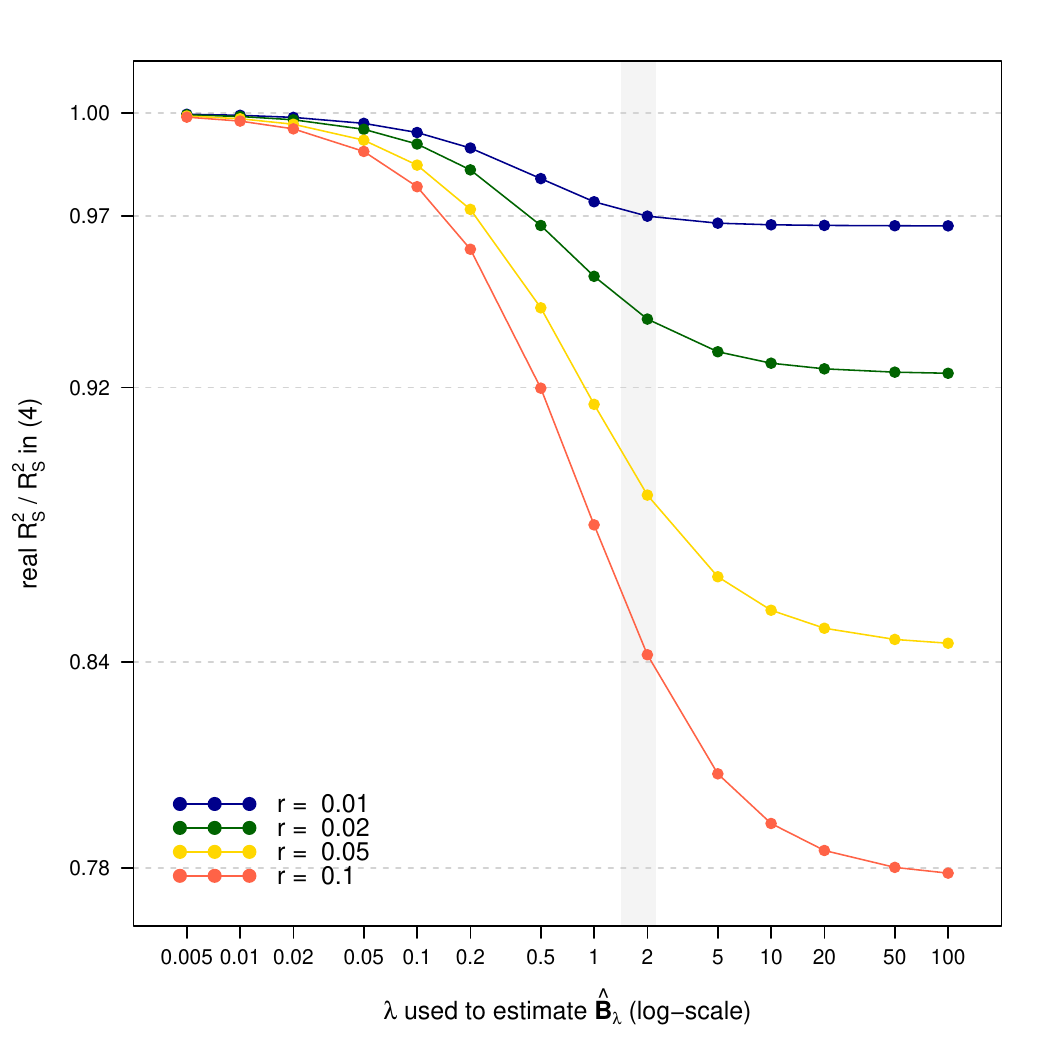}
  \caption{Bias introduced by penalised regression in $\rs(\balpha, \bbeta)$.
    The shaded area corresponds to the range of optimal values for $\lambda$
    used in computing $\BRD$.}
  \label{fig:bias}
\end{figure}

\begin{example}
Consider three predictors $\Xm = \{ X_1, X_2, X_3 \}$ and three sensitive
attributes $\Sm = \{ S_1, S_2, S_3 \}$ distributed as
\begin{equation*}
  \begin{bmatrix} X_1 \\ X_2 \\ X_3 \\ S_1 \\ S_2 \\ S_3 \end{bmatrix} \sim
  N\left(
    \begin{bmatrix} 0 \\ 0 \\ 0 \\ 0 \\ 0 \\ 0 \end{bmatrix},
    \begin{bmatrix}
        1 & 0.3 & 0.3 & 0.3 & 0.3 & 0.3 \\
      0.3 &   1 & 0.3 & 0.3 & 0.3 & 0.3 \\
      0.3 & 0.3 &   1 & 0.3 & 0.3 & 0.3 \\
      0.3 & 0.3 & 0.3 &   1 & 0.3 & 0.3 \\
      0.3 & 0.3 & 0.3 & 0.3 &   1 & 0.3 \\
      0.3 & 0.3 & 0.3 & 0.3 & 0.3 &   1 \\
    \end{bmatrix}
  \right),
\end{equation*}
giving the response variable $\y$ by way of the linear model
\begin{equation*}
  \y = 2X_1 + 3X_2 + 4X_3 + 5S_1 + 6S_2 + 7S_3 + \be
\end{equation*}
with with independent and identically distributed errors $\varepsilon_i \sim
N(0, 100)$. Hence $\bbeta = [2, 3, 4]^\T$ and $\balpha = [5, 6, 7]^\T$ using the
notation established in \mref{eq:komiyama-main}.

Figure \ref{fig:bias} shows the ratio between
\begin{align*}
  \frac{\VAR(\Sm\NCLM{\balpha})}
       {\parbox{0.66\linewidth}{
         $\VAR(\Sm\NCLM{\balpha}) + \VAR(\tU\NCLM{\bbeta}) - \\
         \hspace*{1em} 2\COV(\Sm\NCLM{\balpha}, \tU\NCLM{\bbeta})$}}
\end{align*}
and
\begin{align*}
  \frac{\VAR(\Sm\NCLM{\balpha})}
       {\VAR(\Sm\NCLM{\balpha}) + \VAR(\tU\NCLM{\bbeta})}
\end{align*}
  for various values of the penalisation coefficient
$\lambda$ and $r = {0.01, 0.02, 0.05, 0.10}$. The shaded area represents the
range of the optimal $\lambda$s for the various $\tu$, chosen as those within
1 standard error from the minimum in 10-fold cross-validation as suggested in
\citet{elemstatlearn}. The relative difference between the two is between 3\%
and 5\% for small values like $r = 0.01, 0.02$, and it can grow as large as
16\% for $r = 0.10$. Note that variables are only weakly correlated in this
example; higher degrees of collinearity will result in even stronger bias.

\label{ex:linear}
\end{example}

Using ridge regression to estimate \mref{eq:prep} can be motivated by the need
to address collinearity in $\Sm$, which would make $\BOLS$ numerically unstable
or impossible to estimate. As an alternative, we can replace $\Sm$ with a
lower-dimensional, full-rank approximation based on a reduced number of
principal components in both \mref{eq:prep} and \mref{eq:gamma}. This satisfies
the assumption that $\VAR(\Sm)$ is full rank in the process of deriving
\mref{eq:gamma}.

\section{An Alternative Penalised Regression Approach}
\label{sec:frrm}

\begin{figure}[t]
  \includegraphics[width=\linewidth]{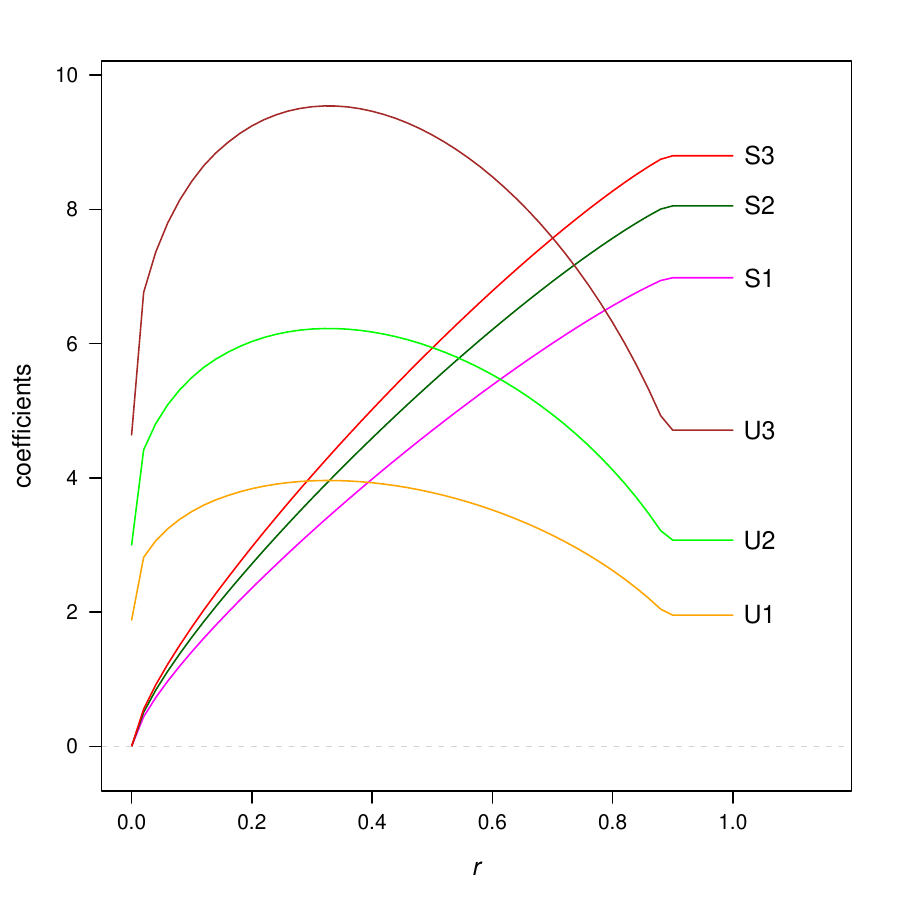}
  \caption{Profile plot for the coefficients estimated by NLCM as a function of
    the bound $r$ for the example in Section \ref{sec:frrm}.}
  \label{fig:nclm}
\end{figure}

The approach proposed by \citet{komiyama} has four limitations that motivated
our work.
\begin{enumerate}
  \item The dimension of the optimisation problem that is solved numerically in
    \mref{eq:gamma} scales linearly in the number of variables.
  \item The formulation of \mref{eq:gamma} allows us to use numeric solvers that
    can handle quadratic programming with quadratic constraints, but cannot be
    translated to regression models other than a linear regression.
  \item The second constraint in \mref{eq:gamma} is undefined in the limit case
    $r = 0$ and can potentially make $(\NCLM{\balpha}, \NCLM{\bbeta})$
    numerically unstable as $r \to 0$.
  \item The behaviour of the estimated regression coefficients is not intuitive
    to interpret. The constraints in \mref{eq:gamma} are functions of both
    $\balpha$ and $\bbeta$: as a result, $\NCLM{\balpha}$ and $\NCLM{\bbeta}$
    are not independent as they would be in an unconstrained OLS regression
    (because $\Sm$ and $\wU$ are orthogonal). Changing the value of the bound
    $r$ then affects the coefficients $\NCLM{\bbeta}$, shrinking or inflating
    them, as well as the $\NCLM{\balpha}$.
\end{enumerate}

\begin{continued}
Consider again the example from Section \ref{sec:bias}. The estimates of the
regression coefficients given by NCLM over $r \in [0, 1]$ are shown in the
profile plot in Figure \ref{fig:nclm}. As expected, we can see that we have all
$\NCLM{\balpha}$ converge to zero as $r \to 0$ because
$\NCLM{\balpha}^\T \VAR(\Sm) \NCLM{\balpha} \to 0$. For $r = 0$, we can say that
$\NCLM{\balpha} = \Zm$ for continuity. As $r$ increases, all $\NCLM{\balpha}$
gradually increase in magnitude. The constraint becomes inactive when
$\rs(\OLS{\balpha}, \OLS{\bbeta}) < r$, hence NCLM reverts back to a standard
OLS regression model for large values of $r$. As a result, all coefficients
stop changing once $r \geqslant 0.85$.

The behaviour of the $\NCLM{\bbeta}$ is, however, difficult to explain on an
intuitive level. They do not change monotonically as $r$ increases, unlike what
happens, for instance, in ridge regression \citep{ridge} or the LASSO
\citep{lasso}. They first increase above $\OLS{\bbeta}$, which helps in reducing
$\rs(\NCLM{\balpha}, \NCLM{\bbeta})$ by increasing its denominator. They then
plateau around $r \approx 0.3$, and start decreasing as
$\rs(\NCLM{\balpha}, \NCLM{\bbeta})$ is allowed to grow.
\end{continued}

\subsection{Fairness by Ridge Penalty}
\label{sec:fropt}

In order to overcome the issues of NCLM we just discussed, we propose an
alternative constrained optimisation framework we call the \textit{fair ridge
regression model} (FRRM). The key idea behind our proposal is to solve the
constrained optimisation problem stated in \mref{eq:general} by imposing a ridge
penalty on $\balpha$ while leaving $\bbeta$ unconstrained. Formally,
\begin{equation}
  \begin{aligned}
    &\min_{\balpha, \bbeta} &
          & \E\left((\y - \wy)^2\right) \\
    &\st  & &\|\balpha\|_2^2 \leqslant t(r)
  \end{aligned}
\label{eq:frrm}
\end{equation}
where $t(r) \geqslant 0$ is such that $\rs(\balpha, \bbeta) \leqslant r$ by
bounding $\aVa$ through $\|\balpha\|_2^2$. Equivalently, we can write
\mref{eq:frrm} as
\begin{equation}
  (\FRRM{\balpha}, \FRRM{\bbeta}) = \argmin_{\balpha, \bbeta}
    \| \y - \Sm\balpha - \wU\bbeta \|_2^2 + \lr \|\balpha\|_2^2
\label{eq:frrm2}
\end{equation}
where $\lr \geqslant 0$ is the value of the ridge penalty that makes
$\rs(\balpha, \bbeta) \leqslant r$. There is a one-to-one relationship between
the values of $t(r)$ and $\lr$, so we choose to focus on the latter. As $r \to
0$, $\lr$ should diverge so that all $\FRRM{\balpha}$ converge to zero
asymptotically and $\FRRM{\balpha}^\T \VAR(\Sm) \FRRM{\balpha} \to 0$ as in
NCLM. Note that zero is a valid value for $r$ in \mref{eq:frrm} while it is not
for NCLM in \mref{eq:gamma}. Furthermore, note that \mref{eq:frrm} is not
specifically tied to $\rs(\balpha, \bbeta)$ (we will show how to replace it with
different fairness constraints in Section \ref{sec:definitions}) and that it can
be easily reformulated with other penalties (which we will discuss in Section
\ref{sec:penalties}).

The $\FRRM{\bbeta}$ are now independent from the $\FRRM{\balpha}$ because the
ridge penalty does not involve the former. Starting from the classical estimator
for the coefficients of a ridge regression (as it can be found in
\citet{ridgenotes} among others), and taking into account that $\Sm$ and $\wU$
are orthogonal, it is easy to show that
\begin{align}
  \begin{bmatrix} \FRRM{\balpha} \\ \FRRM{\bbeta} \end{bmatrix}
  &= \left(
    \begin{bmatrix} \Sm^\T \\ \wU^\T \end{bmatrix}
    \begin{bmatrix} \Sm & \wU \end{bmatrix} +
    \begin{bmatrix}
      \lr \I & \Zm \\
      \Zm    & \Zm
    \end{bmatrix}
    \right)^{-1}
    \begin{bmatrix} \Sm^\T \\ \wU^\T \end{bmatrix} \y \notag \\
  &= \begin{bmatrix}
       \Sm^\T \Sm + \lr \I & \Zm \\
       \Zm                 & \wU^\T \wU
    \end{bmatrix}^{-1}
    \begin{bmatrix} \Sm^\T \\ \wU^\T \end{bmatrix} \y \notag \\
  &= \begin{bmatrix}
       \left(\Sm^\T \Sm + \lr \I\right)^{-1} \Sm^\T \y \\
       (\wU^\T \wU)^{-1} \wU^\T \y
    \end{bmatrix}.
\label{eq:frrm-est}
\end{align}
The $\FRRM{\bbeta}$ can be estimated in closed form, only depend on $\wU$, and
do not change as $r$ varies. The $\FRRM{\balpha}$ depend on $\Sm$ and also on
$r$ through $\lr$, and they must be estimated numerically. However, the form of
$\FRRM{\balpha}$ in \mref{eq:frrm-est} makes it possible to reduce the
dimensionality and the complexity of the numeric optimisation compared to NCLM.
We can estimate them as follows:

\begin{enumerate}
  \item Apply \mref{eq:prep} to $\Sm, \Xm$ to obtain $\Sm, \wU$.
  \item Estimate $\FRRM{\bbeta} = (\wU^\T \wU)^{-1} \wU^\T \y$.
  \item Estimate $\OLS{\balpha} = (\Sm^\T\Sm)^{-1} \Sm^\T \y$. Then:
    \begin{enumerate}
      \item If $\rs(\OLS{\balpha}, \OLS{\bbeta}) \leqslant r$,
        set $\FRRM{\balpha} = \OLS{\balpha}$.
      \item Otherwise, find the value of $\lr$ that satisfies
        \begin{equation}
          \aVa = \frac{r}{1 - r} \FRRM{\bbeta}^\T \VAR(\wU) \FRRM{\bbeta}
        \label{eq:bound2}
        \end{equation}
        and estimate the associated $\FRRM{\balpha}$ in the process.
    \end{enumerate}
\end{enumerate}

As far as determining the value of $\lr$ that results in $\rs(\FRRM{\balpha},
\FRRM{\bbeta}) \leqslant r$ is concerned, we can treat $\FRRM{\bbeta}$ as a
constant that can be pre-computed from $\U$ independently of $\Sm$ and $r$.
Furthermore, $\FRRM{\balpha}$ is available as a closed-form function of $r$
through $\lr$; and we know that $\FRRM{\balpha} \VAR(\Sm) \FRRM{\balpha}^\T
\to 0$ monotonically as $\lr \to \infty$ from the fundamental properties of
ridge regression. As a result, \mref{eq:frrm} is guaranteed to have a single
solution in $\lr$ which can be found with a simple, univariate root-finding
algorithm. Selecting $\lr$ can be though of as model selection, since $\lr$ is a
tuning parameter that affects the distribution of the $\FRRM{\balpha}$.
Estimating the $\FRRM{\balpha}$ given $\lr$ is then a separate model selection
phase.

In the particular case that $\rs(\OLS{\balpha}, \OLS{\bbeta}) \leqslant r$, a
trivial solution to \mref{eq:bound2} is to set $\lr = 0$ and thus
$\FRRM{\balpha} = \OLS{\balpha}$: if the constraint is inactive because the
bound we set is larger than the proportion of the overall variance that is
attributable to the sensitive attributes, then the OLS estimate of $\bbeta$
minimises the objective. This agrees with the behaviour of NCLM shown in Figure
\ref{fig:nclm}. On the other hand, if the constraint is active the objective is
minimised when $\aVa$ takes is largest admissible value, which implies
$\rs(\balpha, \bbeta) = r$. Rewriting \mref{eq:bound} as an equality and moving
all known terms to the right hand gives \mref{eq:bound2}.

In the general case, the ridge penalty parameter is defined on $\mathbb{R}^+$.
However, in \mref{eq:bound2} we can bound it above and below using the equality.
From \citet{lipovetsky}, we have that in a ridge regression with parameter
$\lambda$
\begin{equation*}
  \aVa =
    \y^\T \Sm \A
    \diag\left( \frac{l_i + 2\lr}{(l_i + \lr)^2}\right)
    \A^\T \Sm^\T \y
\end{equation*}
where $\A\LLambda\A^\T = \A\diag(l_i)\A^\T$ is again the eigenvalue decomposition
of $\VAR(\Sm)$. If we replace all the $l_i$ on the right-hand side with the
smallest (respectively, the largest) eigenvalue, we can bound $\aVa$ in
\begin{equation*}
  \left[
    \frac{l_\mathrm{min} + 2\lr}{(l_\mathrm{min} + \lr)^2} \, d,
    \frac{l_\mathrm{max} + 2\lr}{(l_\mathrm{max} + \lr)^2} \, d
  \right]
\end{equation*}
where $d = \y^\T \Sm \A \A^\T \Sm^\T \y$.
We can then replace the bounds above and solve \mref{eq:bound2} as an equality
in $\lr$ to obtain upper and lower bounds for the ridge penalty parameter. If we
let $c = \frac{r}{1 - r} \FRRM{\bbeta}^\T \VAR(\wU) \FRRM{\bbeta}$, the
resulting equations are
\begin{align*}
  &\frac{l_\mathrm{min} + 2\lr}{(l_\mathrm{min} + \lr)^2} \, c = d&
  &\text{and}&
  &\frac{l_\mathrm{max} + 2\lr}{(l_\mathrm{max} + \lr)^2} \, c = d
\end{align*}
which are quadratic equations with one positive solution each. (Clearly, the
respective negative solutions are not admissible since $\lr \geqslant 0$.)

\begin{figure}[t]
  \includegraphics[width=\linewidth]{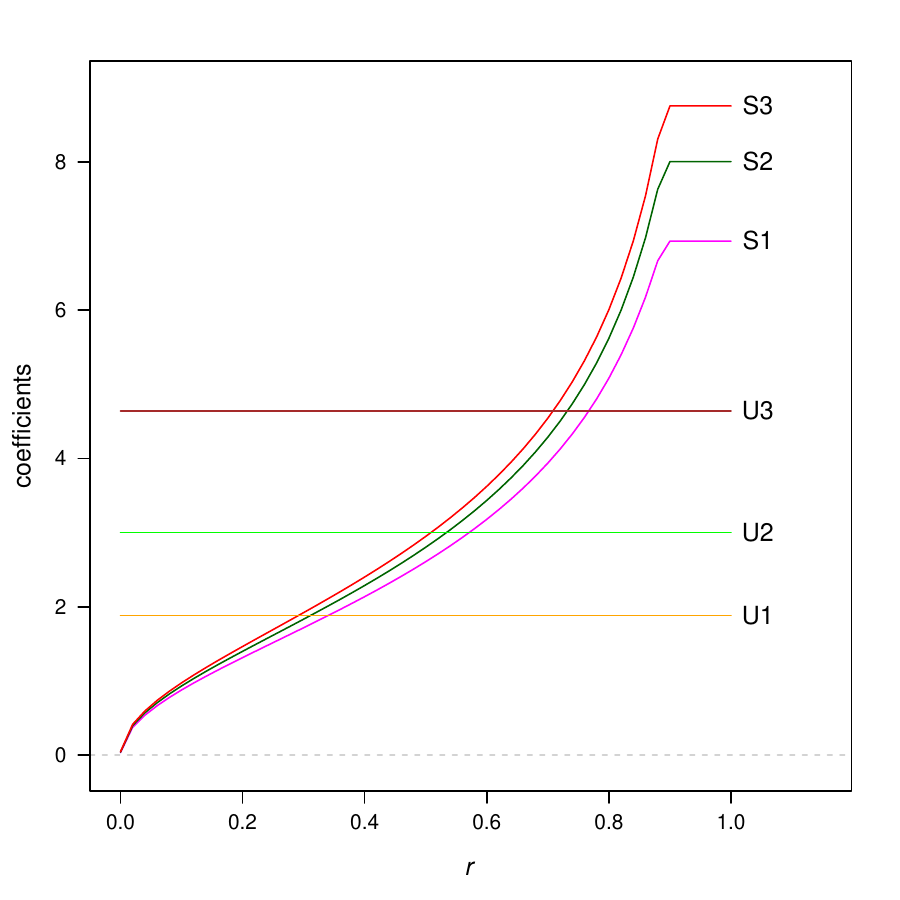}
  \caption{Profile plot for the coefficients estimated by FRRM as a function of
    the bound $r$ for the example in Section \ref{sec:frrm}.}
  \label{fig:frrm}
\end{figure}

\begin{continued}
Consider one more time our example: the regression coefficients
$(\FRRM{\balpha}, \FRRM{\bbeta})$ are shown in Figure \ref{fig:frrm}. The
regression coefficients $\FRRM{\balpha}$ for $S_1$, $S_2$ and $S_3$ still
converge to zero as $r \to 0$, ensuring that $\FRRM{\balpha} \VAR(\Sm)
\FRRM{\balpha}^\T \to 0$, as in Figure \ref{fig:nclm}. However, the
coefficients $\FRRM{\bbeta}$ for $X_1$, $X_2$ and $X_3$ do not change as $r$
changes: they are equal to their OLS estimates for all values of $r$ as implied
by \mref{eq:frrm-est}.
\end{continued}

\subsection{Analytical Solution for Independent $\Sm$}
\label{sec:analytical}

In some instances, it is possible to solve \mref{eq:frrm2} exactly and in closed
form instead of relying on numerical optimisation. This allows us to better
explore the behaviour of $\lambda(r)$ and $\FRRM{\balpha}$, as well as the
effect of relaxing the constraint of statistical parity.

Assume that $\Sm$ is a $q \times n$ matrix of sensitive attributes which are
mutually independent, that is, each $S_j$ is independent of $S_i$ for all $i, j
= 1, \ldots, q, i \neq j$. Furthermore, assume that each $S_j$ is scaled to
$\VAR(S_j) = 1$ in addition to being centred. Let $\Xm$, instead, be a $p \times
n$ matrix of predictors which are allowed to be correlated. Then, it is possible
to write the solution of \mref{eq:bound2} with respect to $\lambda$ in closed
form.

Let $c$ be defined as $c^2:=\FRRM{\bbeta}^\T \VAR(\wU) \FRRM{\bbeta}$ (note that
this is slightly different from the previous definition of c). Then
\mref{eq:bound2} becomes
\begin{align*}
  \left(\frac{1}{n + \lambda} \II_q \Sm^\T\y\right)^\T
  \II_q
  \left(\frac{1}{n + \lambda} \II_q \Sm^\T\y\right) = \frac{r}{1 - r} c^2.
\end{align*}
using the fact that $\Sm^\T\Sm = n \II_q$, where $\II_q$ is the identity matrix
of size $q$. Solving the matrix products we get
\begin{align*}
  \sum_{j = 1}^q (S_j^\T\y)^2 - \frac{r}{1-r} c^2 (n + \lambda)^2 = 0,
\end{align*}
which has solution
\begin{equation}
  \lambda(r) = -n + \frac{\|\Sm^\T\y\|_2^2}{c \sqrt{r/(1 - r)}}
\label{eq:lambda}
\end{equation}
where $\|\Sm^\T\y\|_2^2 = \sum_{j = 1}^q (S_j^\T\y)^2$ is the squared Euclidean
norm in $\mathbb{R}^n$. Plugging \mref{eq:lambda} into \mref{eq:bound2} gives
\begin{equation}
  \FRRM{\balpha} =
    c \sqrt{\frac{r}{1 - r}} \frac{\Sm^\T\y}{\|\Sm^\T\y\|_2^2}.
\label{eq:alpha_FRRM}
\end{equation}
From \mref{eq:alpha_FRRM} we see that the $\FRRM{\balpha}$ increase in magnitude
(that is, move away from zero) as a function of $r$.

As we noted in Section \ref{sec:fropt}, $r = 0$ satisfies statistical parity
exactly: $\FRRM{\balpha} = 0$ and therefore $\wy$ is independent of $\Sm$. As
$r$ increases, $\FRRM{\balpha}$ grows and so does the correlation between $\wy$
and $\Sm$:
\begin{align*}
  \COR(\wy, \Sm)
    &= \frac{\COV(\FRRM{\balpha} \Sm + \FRRM{\bbeta}\wU, \Sm)}
            {\sqrt{\VAR(\wy) \II_q}}\\
    &= \frac{\FRRM{\balpha}}
            {\sqrt{\FRRM{\balpha}^2\VAR(\Sm)+\FRRM{\bbeta}^2\VAR(\wU)}},
\end{align*}
which is again proportional to $\sqrt{r/(1 - r)}$ and equal to $0$ when
$r = 0$.

If the sensitive attributes are not mutually independent, solving
\mref{eq:bound2} exactly is not possible in general and we revert to the
root-finding algorithm described in Section \ref{sec:frrm}. Even with just two
sensitive attributes $\Sm = [S_1, S_2]$, the right hand side of \mref{eq:bound2}
becomes:
\begin{align*}
  &\begin{bmatrix}
  A\\
  B
  \end{bmatrix}^\T
  \begin{bmatrix}
  n + \lambda & C \\
  C & n + \lambda
  \end{bmatrix}^{-1}
  \begin{bmatrix}
  1 & \frac{C}{n} \\
  \frac{C}{n} & 1
  \end{bmatrix}
  \begin{bmatrix}
  n + \lambda & C \\
  C & n + \lambda
  \end{bmatrix}^{-1}
  \begin{bmatrix}
  A\\
  B
  \end{bmatrix} \\
  &\quad = \frac{(A(n + \lambda) - BC)^2+ (B(n + \lambda) - AC)^2}
          {((n + \lambda)^2 - C^2)^2} \\
  &\qquad +\frac{2C/n(A(n + \lambda) - BC)(B(n + \lambda) - AC)}
               {((n + \lambda)^2 - C^2)^2} \\
  &\quad = \frac{(n + \lambda)^2\left(A^2 + B^2 + 2ABC/n\right)}
          {((n + \lambda)^2 - C^2)^2} \\
  &\qquad -\frac{2C(n + \lambda)\left(2AB + CA^2/n + CB^2/n\right)}
               {((n + \lambda)^2 - C^2)^2} \\
  &\qquad +\frac{C^2 \left(A^2 + B^2 + 2ABC/n\right)}
               {((n + \lambda)^2 - C^2)^2} .
\end{align*}
where $A = S_1^\T \y, B = S_2^\T \y,C = S_1^\T S_2 = S_2^\T S_1$ for brevity.
Equating the expression above to $c^2 r/(1 - r)$ will give a 4th-degree
polynomial in $\lambda$. The solutions to this equation can be computed exactly,
but the resulting expression for $\lambda$ do not provide any immediate
insights.

\section{Possible Extensions}
\label{sec:extensions}

FRRM has a simple and modular construction that can accommodate a wide range of
extensions: some examples are modelling nonlinear relationships, incorporating
different and more complex penalties, using different definition of fairness
and handling different types of responses with generalised linear models. The
separation between model selection (the choice of $\lr$) and model estimation
(estimating $\FRRM{\balpha}$ and $\FRRM{\bbeta}$) makes it possible to change
how either or both are performing drawing extensively from established
statistical literature.

\subsection{Nonlinear Regression Models}

We can incorporate kernels into FRRM by fitting the model in the transformed
feature spaces $Z_\Sm(\Sm)$ and $Z_{\wU}(\wU)$ produced by some positive kernel
function, as in \citet{komiyama}. Combining the kernel trick with a ridge
penalty produces a kernel ridge regression model \citep{krr}, which can be
estimated efficiently following \citet{krr2}. Furthermore, this approach
suggests further extensions to Gaussian process regressions, since the two
models are closely related as discussed in \citet{kanagawa}.

\subsection{Different Penalties}
\label{sec:penalties}

We may also want to regularise the $\bbeta$ coefficients to improve predictive
accuracy and to address any collinearity present in the data. One option is to
add a ridge penalty to the $\bbeta$ in addition to that on the $\balpha$.
Ideally, without making $\FRRM{\balpha}$ and $\FRRM{\bbeta}$ dependent to
preserve the intuitive behaviour of the regression coefficient estimates
produced by FRRM. A simple way is to add a second penalty term to
\mref{eq:frrm2},
\begin{multline*}
  (\FRRM{\balpha}, \FRRM{\bbeta}) = \\
    \argmin_{\balpha, \bbeta} \| \y - \Sm\balpha - \Xm\bbeta \|_2^2 +
      \lambda_1(r) \|\balpha\|_2^2 + \lambda_2 \| \bbeta \|_2^2,
\end{multline*}
resulting in
\begin{equation}
  \begin{bmatrix} \FRRM{\balpha} \\ \FRRM{\bbeta} \end{bmatrix}
  = \begin{bmatrix}
       \left(\Sm^\T \Sm + \lambda_1(r) \I\right)^{-1} \Sm^\T \y \\
       \left(\wU^\T \wU + \lambda_2 \I\right)^{-1} \wU^\T \y
    \end{bmatrix}.
\label{eq:doubleridge}
\end{equation}
This is sufficient to ensure there are no unaddressed collinearities as
$\Sm$ and $\wU$ are orthogonal by construction.

\begin{figure}[t]
  \includegraphics[width=\linewidth]{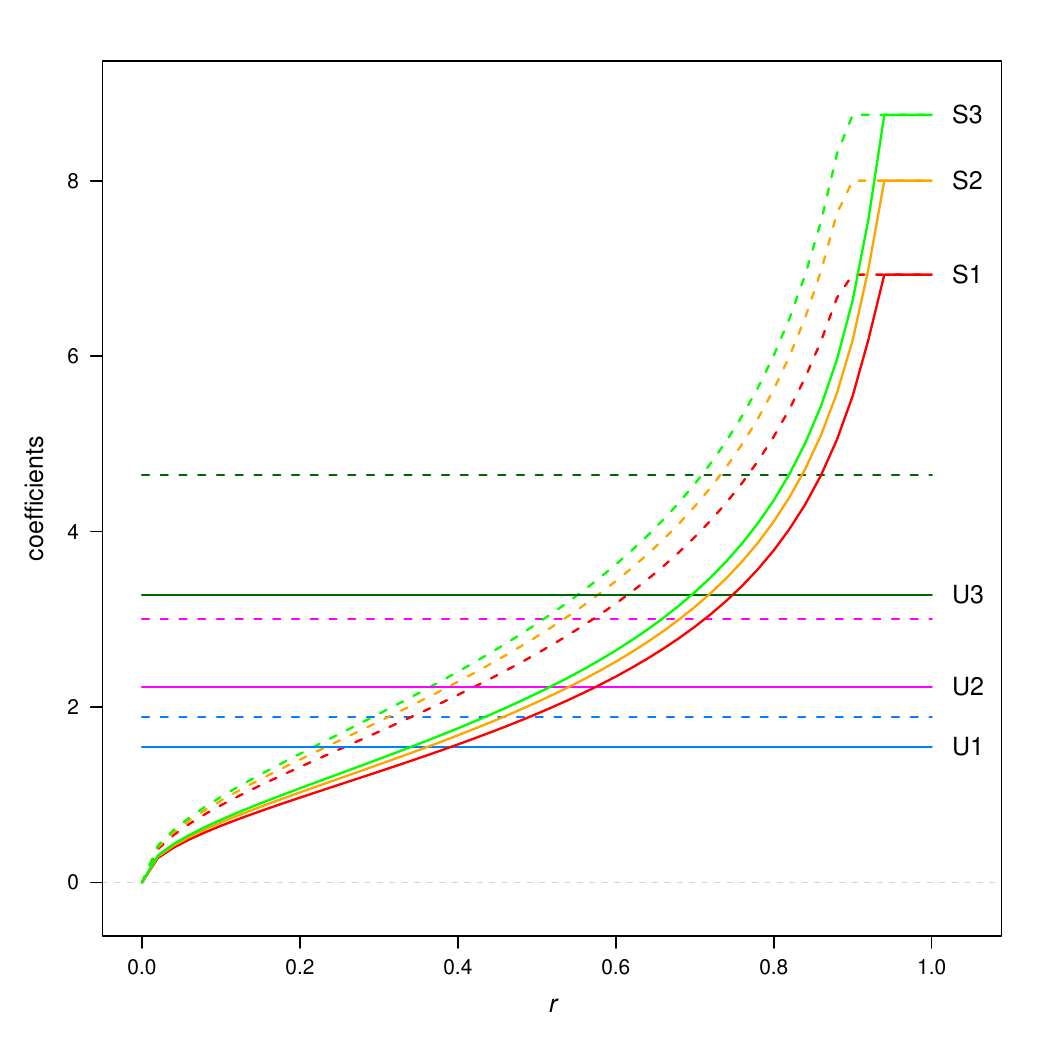}
  \caption{Regression coefficients estimated by FRRM with $\lambda_2 = 10$
    (solid lines) compared to those estimated without penalising the $\bbeta$s
    (dashed lines, reported from Figure \ref{fig:frrm}). Lines in the same
    colour correspond to the same coefficient.}
  \label{fig:showridge}
\end{figure}

\begin{continued}
Figure \ref{fig:showridge} shows the estimates $(\FRRM{\balpha}, \FRRM{\bbeta})$
obtained with $\lambda_2 = 10$ as a function of $r$. If we compare these new
coefficients (solid lines) with those from Figure \ref{fig:frrm}
(dashed lines with the same colours), we can see that the $\FRRM{\bbeta}$
are still independent from $r$. At the same time, they have been shrunk
towards zero and that means that $\FRRM{\bbeta} \VAR(\wU) \FRRM{\bbeta}^\T$ is
also smaller than before. As a result, we need a larger $\lr$ to produce
estimates of $\FRRM{\balpha}$ small enough to satisfy the bound in
\mref{eq:bound}. The $\FRRM{\balpha}$ in Figure \ref{fig:showridge} are smaller
than the corresponding $\FRRM{\balpha}$ in Figure \ref{fig:frrm}.
\end{continued}

It is also interesting to note that \mref{eq:frrm} can be implemented with
penalised models other than a ridge regression. Any model that can shrink the
coefficients associated with the sensitive attributes towards zero, thus
decreasing the proportion of variance they explain in the response, can control
the value of the bound $\rs(\balpha, \bbeta)$ as a function of the tuning
parameter. One possibility is to replace the ridge penalty with a LASSO penalty
in order to perform feature selection on the sensitive attributes, a problem
also investigated in \citet{zafar2}, \citet{kazemi} and \citet{nafea}. Or we can
combine it with the ridge penalty to obtain an elastic net model \citep{enet},
which will often provide better predictive accuracy.

\subsection{Different Definitions of Fairness}
\label{sec:definitions}

The modular approach to fairness used in \mref{eq:frrm} makes it possible to
change the definition of fairness and its implementation in the bound used in
model selection without affecting the estimation of $\balpha$ and $\bbeta$.

For instance, \mref{eq:frrm2} uses $\rs(\balpha, \bbeta) \leqslant r$ as a bound
to enforce fairness as defined by statistical parity. But we can replace it with
a similar bound for equality of opportunity, such as
\begin{equation*}
  \reo(\pphi, \psi) = \frac{\VAR(\Sm\pphi)}{\VAR(\y\psi + \Sm\pphi)}
\end{equation*}
where $\wy$ is defined as before and $\pphi$, $\psi$ are the coefficients of
the regression model
\begin{equation*}
  \wy = \y\psi + \Sm\pphi + \varepsilon^*.
\end{equation*}
If equality of opportunity holds exactly, $\wy$ is independent from $\Sm$ given
$\y$ and $\COV(\wy, \Sm \given \y) = 0$. Then all $\pphi$ are equal to zero and
$\reo(\pphi, \psi) = 0$. FRRM can achieve that asymptotically as $\lr \to
\infty$ because $\wy \to \wU\FRRM{\bbeta}$ and $\wU$ is orthogonal to $\Sm$. If,
on the other hand, $\lr \to 0$ the constraint becomes inactive because the
$\FRRM{\balpha}$ converge to the corresponding $\OLS{\balpha}$. For finite,
positive values of $\lr$, we have that $\wy = \wU\FRRM{\bbeta} +
\Sm\FRRM{\balpha}$ and therefore $|\COV(\wy, \Sm \given \y)|$ will decrease as
$\lr$ increases. This allows us to control $\reo(\pphi, \psi)$ in the same way
as we did $\rs(\balpha, \bbeta)$ in Section \ref{sec:fropt}.

A further advantage of enforcing fairness in this way is that we can control
both statistical parity and equality of opportunity as a function of $r$
(through $\lr$) at the same time. So, for instance, we could replace the
constraint in \mref{eq:frrm} with
$\max\{\rs(\balpha, \bbeta), \reo(\pphi, \psi)\}$ or a convex combination
$w\rs(\balpha, \bbeta) + (1 - w)\reo(\pphi, \psi)$, $w \in (0, 1)$. Few
approaches in the literature combine different definitions of fairness in the
same model; one example is \citet{berk}.

We can also choose to enforce individual fairness. Following along the lines of
\citet{berk}, we can start by defining a penalty function
\begin{equation*}
  f(\balpha, \y, \Sm) =
    \sum\nolimits_{i, j} d(y_i, y_j) (\si\balpha - \sj\balpha)^2
\end{equation*}
that penalises models in which individuals $i$ and $j$ with profiles
\mbox{$(y_i, \ui, \si)$} and \mbox{$(y_j, \uj, \sj)$} receive differential
treatment in proportion to $(\si\balpha - \sj\balpha)^2$. If two individuals
take the same values for the sensitive attributes, $\si = \sj$ and their term
vanish from the sum. If $\si \neq \sj$, the corresponding term increases with
both the difference in the outcomes, measured by some distance $d(y_i, y_j)$,
and with the difference in their sensitive attributes $\si$ and $\sj$.

If $\lr \to \infty$, then $(\si\FRRM{\balpha} - \sj\FRRM{\balpha})^2 \to 0$
because all coefficients in $\FRRM{\balpha}$ are shrunk towards zero. As a
result, $f(\FRRM{\balpha}, \y, \Sm)$ converges to zero as well. On the other
hand, if $\lr \to 0$ then $f(\FRRM{\balpha}, \y, \Sm) \to
f(\OLS{\balpha}, \y, \Sm)$ to take its maximum value.

For consistency with $\rs(\balpha, \bbeta)$ and $\reo(\balpha, \bbeta)$, we then
construct the constraint to use in \mref{eq:frrm} by normalising
$f(\balpha, \y, \Sm)$ as
\begin{equation*}
  \DIF = \frac{f(\FRRM{\balpha}, \y, \Sm)}{f(\OLS{\balpha}, \y, \Sm)}
\end{equation*}
so that the bound $r$ is defined in $[0, 1]$ as before. This is convenient for
interpretation and to include $\DIF$ in a convex combination with other fairness
definitions.

\begin{figure}[t]
  \includegraphics[width=\linewidth]{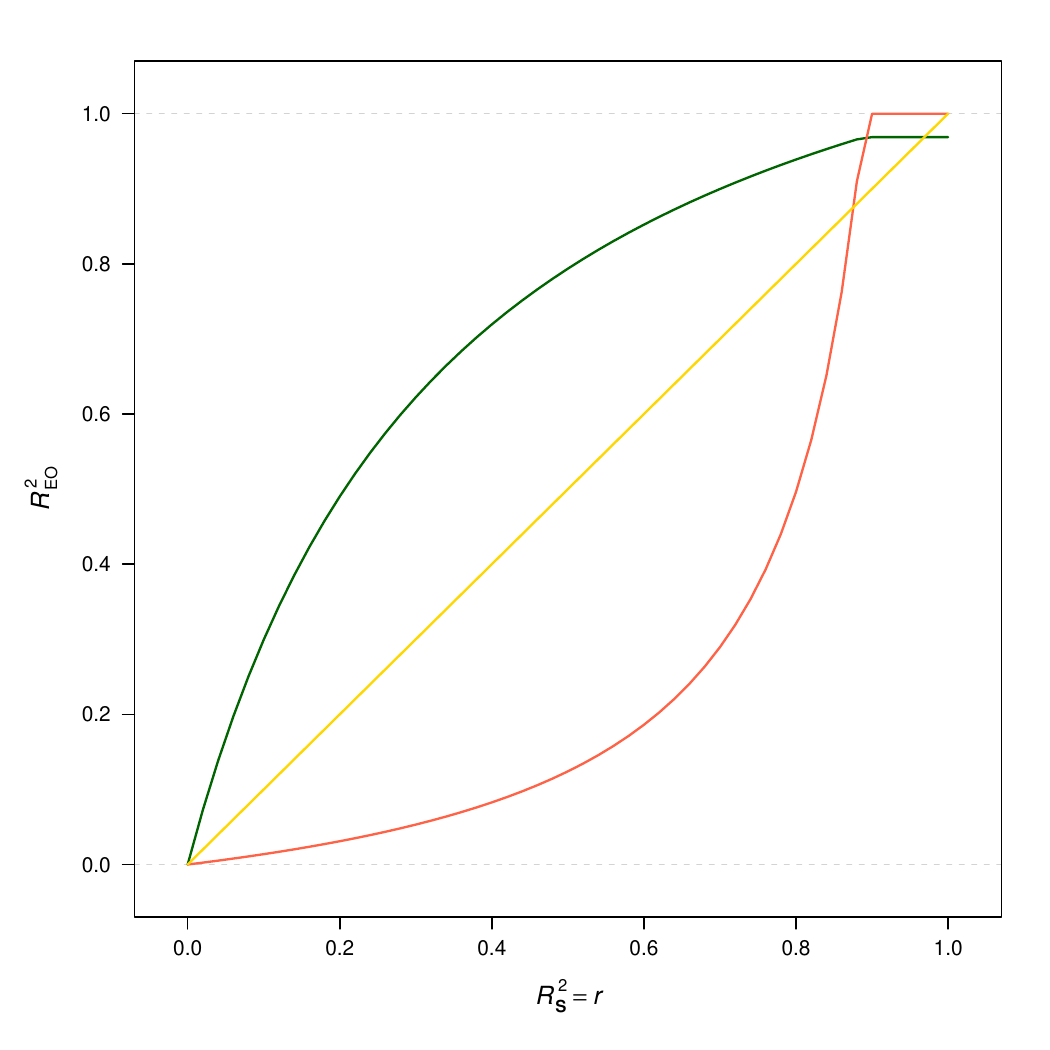}
  \caption{$\reo(\pphi, \psi)$ (green) and $\DIF$ (orange) as a function of
    $\rs(\FRRM{\balpha}, \FRRM{\bbeta}) = r$ (yellow).}
  \label{fig:speo}
\end{figure}

\begin{continued}
Consider the example from Section \ref{sec:bias} one last time. Figure
\ref{fig:speo} shows the estimates of $\reo(\pphi, \psi)$ and $\DIF$ as a
function of $\rs(\FRRM{\balpha}, \FRRM{\bbeta}) = r \in [0, 1]$. For the sake of
the example, we choose $d(y_i, y_i) = |y_i - y_j|$ in $\DIF$. As $r$ increases,
that is, as $\lr \to 0$, all of $\rs(\FRRM{\balpha}, \FRRM{\bbeta})$,
$\reo(\pphi, \psi)$ and $\DIF$ increase monotonically. Hence any function that
preserves their joint monotonicity can be used to enforce a user-specified
combination of statistical parity, equality of opportunity and individual
fairness.
\end{continued}

\subsection{Generalised Linear Models}

Another possible extension of FRRM is to adapt \mref{eq:frrm2} to generalised
linear models \citep[GLMs;][]{glm} including Cox's proportional hazard models
\citep{cox}. This makes it possible to introduce fair modelling in the extensive
range of applications in which GLMs are a \emph{de facto} standard while still
being able to follow the best practices developed in the literature for those
applications (significance testing, model comparison, confidence intervals for
parameters, meta analysis, etc.).

Minimising the sum of squared residuals in a linear regression is a particular
case of minimising the deviance $D(\cdot)$ (that is, $-2$ times the
log-likelihood) of a generalised linear model, which we can constrain to ensure
that we achieve the desired level of fairness. Starting from the general
formulation of a GLM
\begin{align*}
  &\E(\y) = \bmu,&
  &\bmu = g^{-1}(\eeta),&
  &\eeta = \Sm\balpha + \wU\bbeta,
\end{align*}
where $g(\cdot)$ is the link function, we can draw on \citet{glmenet} and
replace \mref{eq:frrm2} with
\begin{equation}
  (\FRRM{\balpha}, \FRRM{\bbeta}) = \argmin_{\balpha, \bbeta}
    D(\balpha, \bbeta) + \lambda(r) \|\balpha\|_2^2.
\label{eq:fgrrm}
\end{equation}
The ridge penalty $\lambda(r)$ can then be estimated to give
\begin{equation}
  \frac{D(\balpha, \bbeta) - D(\Zm, \bbeta)}
       {D(\balpha, \bbeta) - D(\Zm, \Zm)} \leqslant r.
\label{eq:glmbound}
\end{equation}
We call this approach a \emph{fair generalised ridge regression model} (FGRRM).

For a Gaussian GLM, \mref{eq:fgrrm} is identical to \mref{eq:frrm2} because the
deviance is just the residual sum of squares, and \mref{eq:glmbound} simplifies
to $\rs(\balpha, \bbeta) \leqslant r$. For a Binomial GLM with the canonical
link function \mbox{$\eeta = \log(\bmu / (1 - \bmu))$}, that is, a logistic
regression, \mref{eq:glmbound} bounds the difference made by $\Sm\balpha$ in
the classification odds. For a Poisson GLM with the canonical link
\mbox{$\eeta = \log\bmu$}, that is, a log-linear regression, \mref{eq:glmbound}
bounds the difference in the intensity (that is, the expected number of arrivals
per unit of time).

In the case of Cox's proportional hazard model for survival data, we can write
the hazard function as
\begin{equation*}
  h(t; \wU, \Sm) = h_0(t)\exp(\Sm\balpha + \wU\bbeta)
\end{equation*}
where $h_0(t)$ is the baseline hazard at time $t$. The corresponding deviance
can be used as in \mref{eq:fgrrm} and \mref{eq:glmbound} to enforce the desired
level of fairness, bounding the ratio of hazards through the difference in the
effects of the sensitive attributes. The computational details of estimating
this model are described in \citet{coxenet}.

\citet{glmenet} and \citet{coxenet} describe how to fit GLMs and Cox's
proportional hazard models with an elastic net penalty, which is a further
extension to the application of FRRM to this class of models. We may also
consider adapting one of the several pseudo-$R^2$ coefficients available in the
literature, such as \citet{nagelkerke}'s or \citet{tjur}'s, to replace
\mref{eq:glmbound}.

\begin{figure}[b]
  \includegraphics[width=\linewidth]{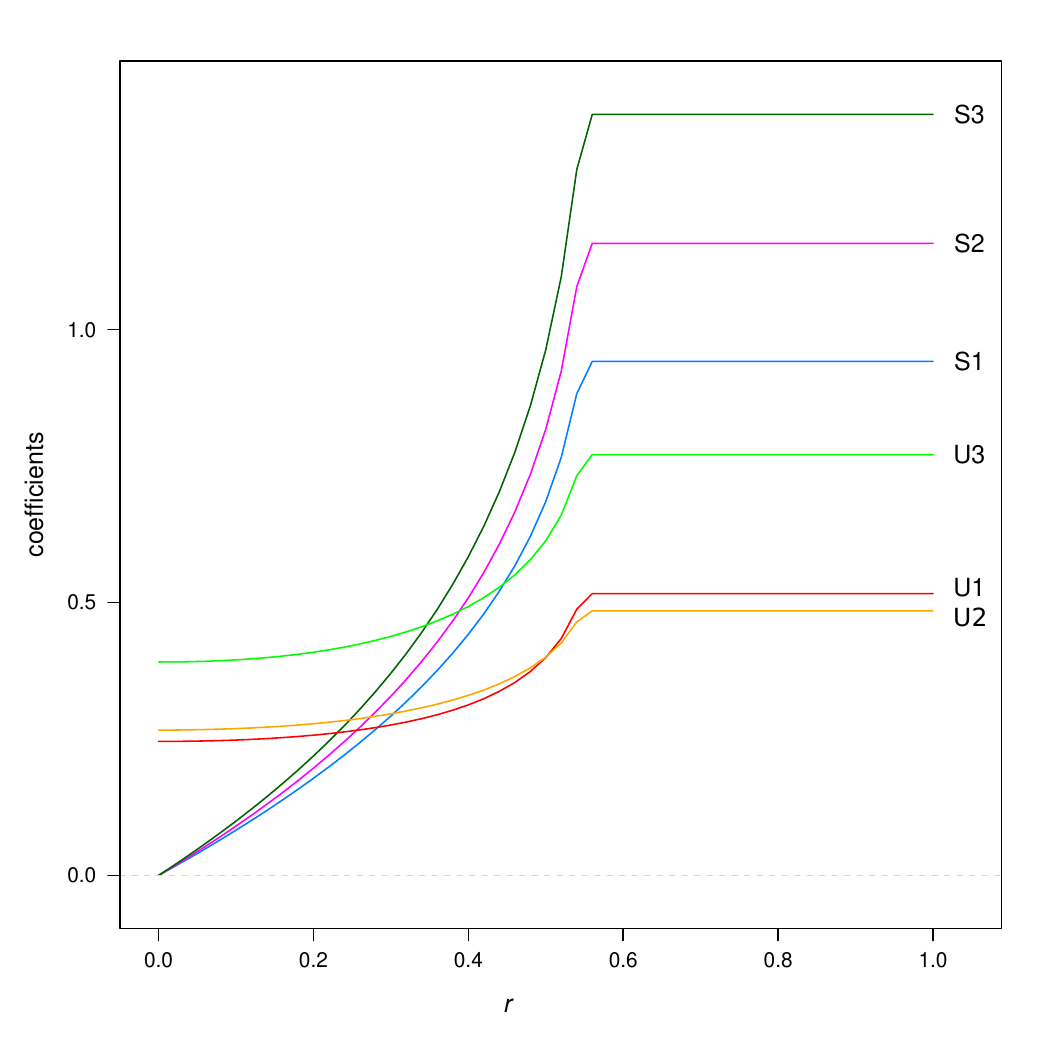}
  \caption{Profile plot for the coefficients estimated by FGRRM as a function of
    the bound $r$ for the model in Example \ref{ex:logistic}.}
  \label{fig:fgrrm}
\end{figure}

Finally, we note that the $\FGRRM{\bbeta}$ are not constant over $r$ in GLMs
with a fixed scale factor (such as logistic and log-linear regressions): their
values depend on the residual deviance, which changes as a function of $r$
through the $\FGRRM{\balpha}$. This phenomenon is described in detail in
\citet{mood}, and we illustrate it with the example below.

\begin{example}
Consider again the $\Xm$ and $\Sm$ from Example \ref{ex:linear}, this time in
the context of a logistic regression with linear component
\begin{equation*}
  \eeta = 1 + 0.5 X_1 + 0.6 X_2 + 0.7 X_3 + 0.8 S_1 + 0.9 S_2 + S_3.
\end{equation*}
The estimates of the regression coefficients given by FGRRM over $r \in [0, 1]$
are shown in Figure \ref{fig:fgrrm}. The $\FGRRM{\balpha}$ are all equal to zero
when $r = 0$, and they gradually increase to reach corresponding $\OLS{\balpha}$
as in Figure \ref{fig:frrm}. In doing that, they gradually explain more and more
of the deviance of the model, which forces the $\FGRRM{\beta}$ to increase as
well. However, they increase monotonically, unlike the $\NCLM{\bbeta}$, with a
speed that matches that of the $\FGRRM{\balpha}$. The change in scale is driven
by the implicit constraint that a standard logistic distribution has a fixed
variance: any increases in the variance explained by the $\FGRRM{\balpha}$ also
affect the variance of the residuals, thus forcing a rescaling of all
coefficients to satisfy the constraint. We find this behaviour more intuitive to
explain than NCLM's because it closely matches that of an unconstrained
logistic regression as described in \citet{mood}. The estimates of the $\bbeta$
produced by the fair logistic regression model proposed by \citet{zafar}, which
we will use in the experimental validation in Section \ref{sec:logisticsim},
change non-monotonically in $r$ like the $\NCLM{\bbeta}$ (figure not shown for
brevity). Note that using a quasi-Binomial GLM would remove the constraint on
the scale factor and thus allow the $\FGRRM{\bbeta}$ to be constant with respect
to $r$.

\label{ex:logistic}
\end{example}

\section{Experimental Evaluation}
\label{sec:simulations}

We evaluate the performance of F(G)RRM using NCLM and the fair regression models
from \citet{zafar} as baselines. We will label the latter as ZLM
(\textit{Zafar's linear model}) and ZLRM (\textit{Zafar's logistic regression
model}) in the following. All six data sets used in this section are available
in the \textit{fairml} R package \citep{fairml}; we refer the reader to its
documentation for further details on each data set including how they have been
preprocessed. F(G)RRM, NCLM and ZL(R)M are also implemented in \textit{fairml}.

We choose ZL(R)M because it is a current, strong baseline (\citet{zafar} shows
that it outperforms four other methods from recent literature) and because it
uses a definition of fairness that is comparable to that in \mref{eq:glmbound}:
\begin{itemize}
  \item ZLRM controls the effect of the sensitive attributes on the response by
    bounding $\zcov$ marginally for each $S_i$;
  \item ZLM equivalently bounds $\zcovy$, since $\weta = \wy$ in a linear
    regression model.
\end{itemize}
If $\zcov = 0$, then $\wy$ is independent from $S_i$, giving statistical parity.
If $\zcov > 0$, then its magnitude controls the proportion of the variance of
$\weta$ explained in the simple regression model of $\weta$ against $S_i$. This
proportion maps to the proportion of explained variance directly in ZLM, and to
the proportion of explained deviance through the link function $g(\cdot)$ in
ZLRM. The key difference between ZL(R)M and F(G)RRM is that ZL(R)M controls the
overall proportion of variance or deviance explained by the sensitive attributes
marginally for each $S_i$, while F(G)RRM controls it jointly for all $S_i$.

Overall, we find that F(G)RRM is at least as good as the best between NCLM and
ZL(R)M in terms of both predictive accuracy and goodness of fit. In particular:
\begin{itemize}
  \item FRRM outperforms NCLM for all but one data set when $r > 0$.
  \item F(G)RRM outperforms ZL(R)M for all considered data sets and for low values
    of $r$, that is, for models that have strong fairness constraints like those
    we may find in practical applications.
\end{itemize}

\subsection{Fair Linear Regression Models}
\label{sec:linearsim}

\begin{figure*}[p]
  \includegraphics[width=0.95\linewidth]{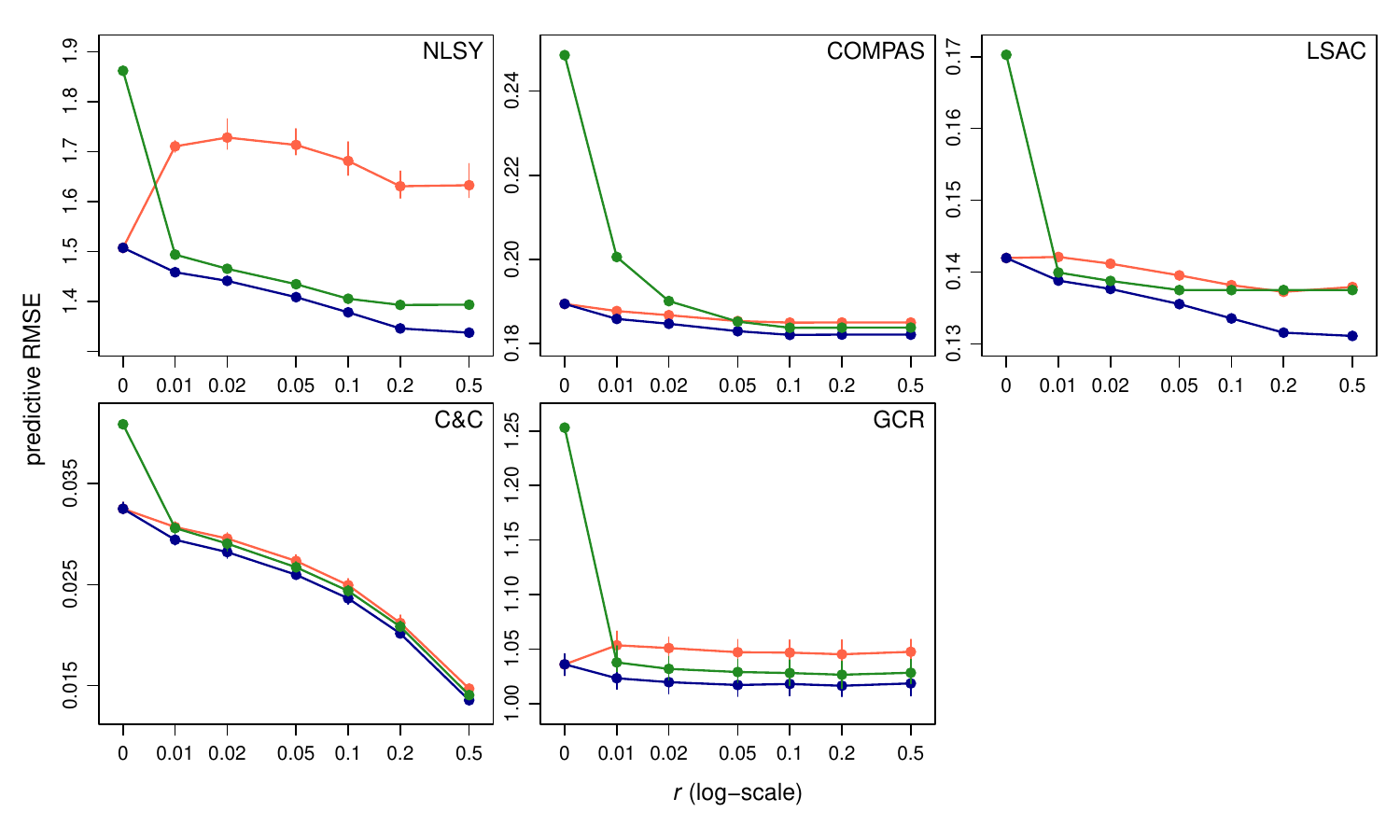}
  \caption{Predictive RMSE for NCLM (orange), FRRM (blue) and ZLM (green) on the
    data sets described in Section \ref{sec:linearsim}. Bars show 90\%
    confidence intervals. Lower values are better.}
  \label{fig:rmse}
  \includegraphics[width=0.95\linewidth]{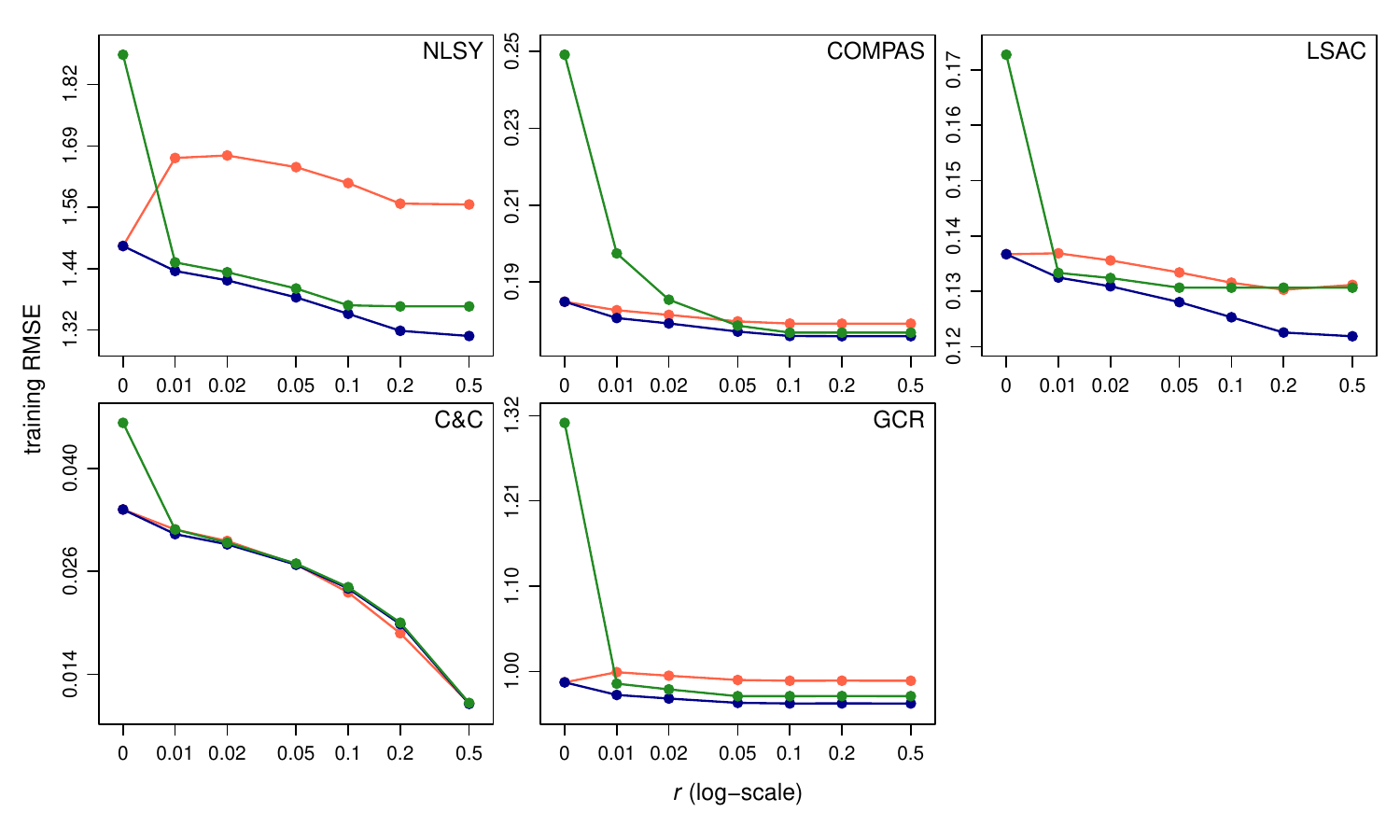}
  \caption{Training RMSE for NCLM (orange), FRRM (blue) and ZLM (green) on the
    data sets described in Section \ref{sec:linearsim}. Bars showing 90\%
    confidence intervals are too small to be visible. Lower values are better.}
  \label{fig:rsquared}
\end{figure*}

We compare FRRM with NCLM and ZLM using the four real-world data sets that were
also used in \citet{komiyama} as well as the German Credit data set \citep{gcr}.
Our results for NCLM differ from those in \citet{komiyama} due to the bias issue
described in Section \ref{sec:bias}, although they do largely agree overall.

The Communities and Crime data set (C\&C, $810$ observations, $101$ predictors)
comprises socio-economic data and crime rates in communities in the United
States: we take the normalised crime rate as the response variable, and the
proportion of African American people and foreign-born people as the sensitive
attributes. The COMPAS data set (COMPAS, $5855$ observations, $13$ predictors)
comprises demographic and criminal records of offenders in Florida: we take
recidivating within two years as the response variable and the offender's gender
and race as the sensitive attributes. The National Longitudinal Survey of Youth
data set (NLSY, $4908$ observations, $13$ predictors) is a collection of
statistics from the U.S. Bureau of Labour Statistics on the labour market
activities and life events of several groups: we take income in 1990 as the
response variable, and gender and age as the sensitive attributes. The Law
School Admissions Council data set (LSAC) is a survey among U.S. law school
students: we take the GPA score of each student as the response variable, and
the race and the age as the sensitive attributes. The German Credit data set
(GCR, $1000$ observations, $42$ predictors) is a collection of $700$ ``good''
loans and $300$ ``bad'' loans with a set of attributes that can be used to
classify them as good or bad credit risks. We take the rate as the response
variable, and the age, gender and foreign-born status as the sensitive
attributes.

\begin{figure*}[t]
  \includegraphics[width=0.95\linewidth]{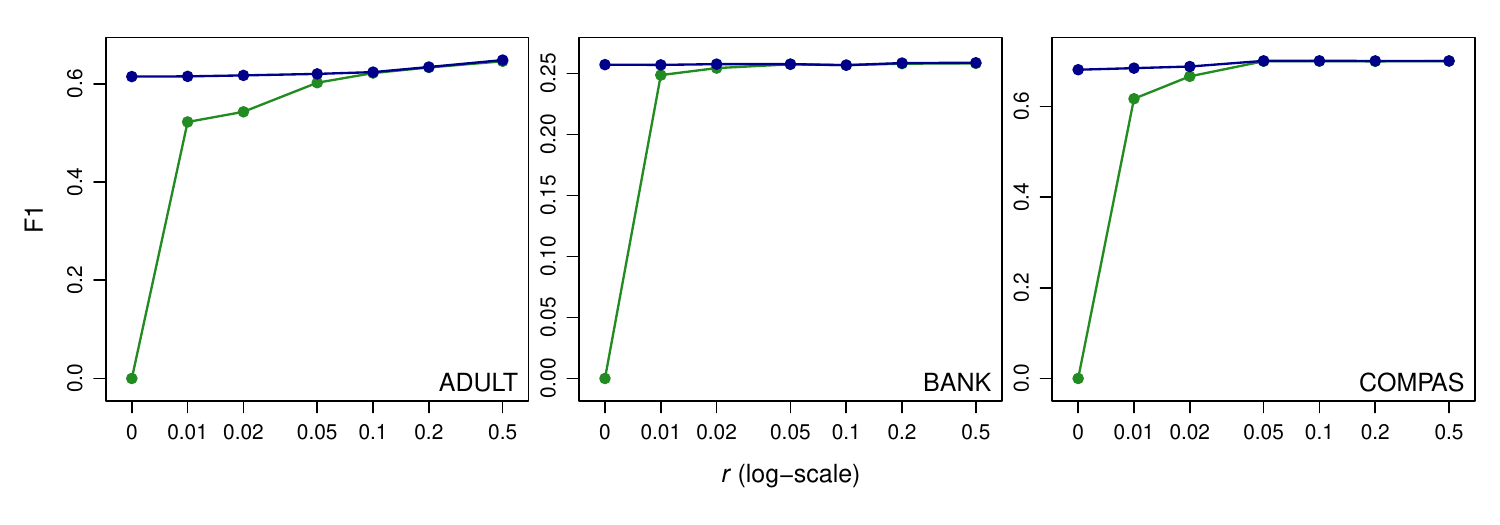}
  \caption{Predictive F1 score for ZLRM (green) and FGRRM (blue) on the data
    sets described in Section \ref{sec:logisticsim}. Bars showing 90\%
    confidence intervals are too small to be visible. Higher values are better.}
  \label{fig:f1}
\end{figure*}

We evaluate both NCLM and FRRM using 50 runs of 10-fold cross-validation with
constraint values $r = \{ 0, 0.01, 0.02, 0.05, 0.10, 0.20, 0.50 \}$. We then
measure the largest resulting $\zcovy $ for each $r$ and we use that as the
bound in ZLM to compare the accuracy of all models for the same level of
fairness. To reduce simulation variability, in each run of cross-validation we
use the same folds for all algorithms. We measure performance with:
\begin{itemize}
  \item the predictive root-mean-square error (RMSE) produced by the model on
    the validation sets in the cross-validation;
  \item the training RMSE produced by the model on the training sets in the
    cross-validation.
\end{itemize}

The predictive RMSE is shown in Figure \ref{fig:rmse}. FRRM consistently
achieves a smaller RMSE than NCLM across all data sets and $r >0$. FRRM also
achieves a smaller RMSE than ZLM in NLSY, COMPAS and LSAC for $r > 0$. In the
case of C\&C and GCR, FRRM achieves a lower RMSE than NCLM and ZLM but the
difference is negligible for practical purposes.

For $r = 0$, FRRM and NCLM estimate the same model containing only the
decorrelated predictors $\wU$ and therefore have the same predictive RMSE. On
the other hand, ZLM has a much higher RMSE than FRRM and NCLM because it
estimates a model that only includes those predictors that are orthogonal to all
sensitive attributes simultaneously ($\zcovy \propto |\COV(\Xm, S_i)| = 0$ for
all $S_i$). However, the empirical covariances between predictors and sensitive
attributes are usually numerically different from zero even when their
theoretical counterparts are not. Hence ZLM ends up dropping more and more
predictors as $r \to 0$ and estimates an intercept-only model for $r = 0$.

The training RMSE is shown in Figure \ref{fig:rsquared}, and follows a similar
pattern to the predictive RMSE in Figure \ref{fig:rmse}. However, it is notable
that both FRRM and ZLM outperform NCLM, despite its theoretical optimality
guarantees, for $r > 0$ in NLSY, LSAC and GCR; and for $r > 0.05$ in COMPAS.
This is possible because the assumptions made in \citet{yamada} and
\citet{komiyama} do not hold. Firstly, \citet{yamada} assume that the constraint
must be active, which is not the case whenever $\rs(\OLS{\balpha}, \OLS{\bbeta})
\geqslant r$. Secondly, both \citet{yamada} and \citet{komiyama} assume that
both $\VAR(\Sm)$ and $\VAR(\wU)$ are full rank. While this is technically true
for all data sets, we note that $\VAR(\Sm)$ has at least one eigenvalue smaller
than $10^{-6}$ in each of COMPAS, LSAC, NLSY and GCR. The fact that FRRM
outperforms NCLM for all these data sets, but not for C\&C, suggests that it is
more numerically robust in this respect; which is expected since ridge
regression does not make any assumption on the nature of $\Sm$.

All algorithms achieve the desired level of fairness on both on the training and
the validation sets in the cross-validation whenever the bound $r$ is active,
that is, when the estimated model does not revert to an OLS regression. In
particular, the level of fairness observed in the predictions for the validation
sets is matches that required when training the models.

\subsection{Fair Logistic Regression Models}
\label{sec:logisticsim}

We now compare FGRRM with ZLRM, following the same steps in Section
\ref{sec:linearsim}. However, we measure both predictive accuracy and goodness
of fit with the F1 score (the harmonic average between precision and recall).

For this purpose, we will use the ADULT and BANK data sets that were also used
in \citet{zafar} as well as COMPAS. The ADULT data set (30162 observations, 14
predictors) contains a set of answers to the U.S. 1994 Census that are relevant
for predicting whether a respondent's income exceeds \$50K. We take the binary
income indicator (whether income is above or below \$50K) as the response
variable, and sex and age as the sensitive attributes. Note that we enforce
fairness for both sensitive attributes simultaneously, while \citet{zafar} only
considered them individually in separate models. The BANK data set (41188
observations, 19 variables) contains information on the phone calls conducted by
a Portuguese banking institution's direct marketing campaigns to convince
prospective clients to subscribe a term deposit. We take the age as the
sensitive attribute and whether the call resulted in a subscription as the
response. The COMPAS data set is the same as in Section \ref{sec:linearsim}, but
we now treat the response variable as a discrete binary variable.

The results for predictive accuracy are shown in Figure \ref{fig:f1}. FGRRM
systematically outperforms ZLRM for $r \leqslant 0.05$ for both ADULT and
COMPAS, and the two models have equivalent performance for $r > 0.1$. In the
first case the bounds are active; in the latter they are not, and both FGRRM
and ZLRM revert back to unconstrained logistic regression models. As for the
BANK data set, FGRRM and ZLRM have equivalent performance for all values of $r$
because BANK contains just one sensitive attribute. Therefore, controlling the
proportion of deviance explained by sensitive attributes marginally (in ZLRM) is
the same as controlling it jointly (in FGRRM). For $r = 0$, ZLRM suffers a
catastrophic loss in predictive accuracy for the same reasons as ZLM.

The observed goodness of fit follows the same patterns as predictive accuracy
for all data sets, save for the fact that the F1 scores are higher by up to
0.02. Furthermore, both FGRRM and ZLM achieve the desired level of fairness as
was the case for the models in Section \ref{sec:linearsim}.

\subsection{Comparison with Other Fair Models in the Literature}
\label{sec:othersim}

The limitations intrinsic to other fair models in the literature prevent us from
comparing them with F(G)RRM as thoroughly as we did for NCLM and ZL(R)M.
Nevertheless, we can draw some limited results to get a partial view of how
their performance relates to that of F(G)RRM. Here we consider the models built
on statistical parity proposed by \citet{steinberg} and \citet{reductions}.

\begin{figure*}[h!]
  \includegraphics[width=0.95\linewidth]{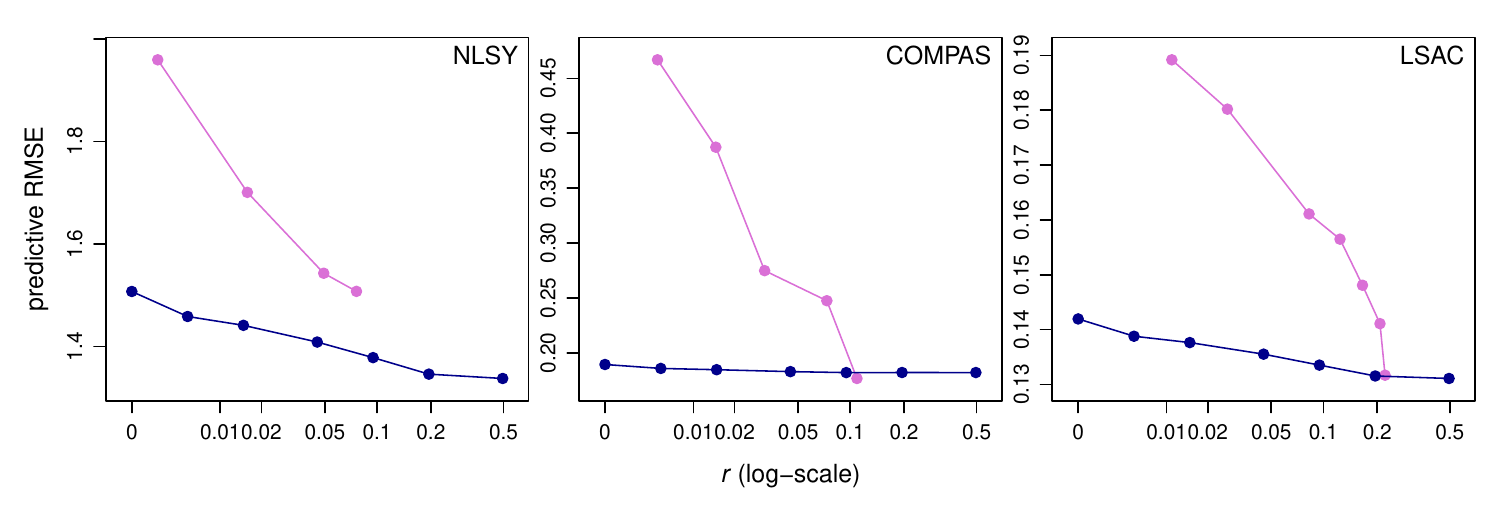}
  \caption{Predictive RMSE for FRRM (blue) and the approach from
    \citet[][violet]{steinberg} on the data sets to which both apply. Bars
    showing 90\% confidence intervals are too small to be visible. Lower values
    are better.}
  \label{fig:steinberg}
\end{figure*}

The fair regression model proposed by \citet{steinberg} uses an auxiliary
logistic regression model to control the effect of a single binary sensitive
attribute on $\y$. The optimal regression is chosen as the model that maximises
a penalised loglikelihood score computed as follows for a given penalty
$\gamma$:
\begin{itemize}
  \item They estimate the main regression model \\
    $\y = \Sm\bbeta_{\Sm} + \Xm\bbeta_{\Xm}$ to obtain $\wy$.
  \item They approximate an auxiliary logistic regression of $\Sm$ on $\wy$ and
    then approximate the mutual information between $\Sm$ and its fitted values
    $\widehat{\Sm}$.
  \item They add a penalty term equal to $\gamma$ times the mutual information
    above to the loglikelihood of the main regression model to promote fairness.
\end{itemize}

\citet{steinberg} do not provide an implementation of their proposed linear
regression model. In our own implementation, we extend it in two ways to allow
for a meaningful comparison with FRRM:
\begin{itemize}
  \item we allow more than one binary sensitive attribute in the model by adding
    a separate penalty term for each, all with the same coefficient $\gamma$;
  \item we allow sensitive attributes with more than two values by using a
    multinomial logistic regression as the auxiliary model.
\end{itemize}
Even so, we are limited to the COMPAS data (same sensitive attributes as
before), the NLSY data (gender as the only sensitive attribute) and the LSAC
data (race as the only sensitive attribute). Furthermore, we are unable to
control $r$ exactly due to the highly nonlinear relationship between $\gamma$
and $r$. The predictive RMSE for FRRM and for the model from \citet{steinberg}
are shown in Figure \ref{fig:steinberg}: the former dominates the latter for $r
< 0.1$ for all three data sets. The right-most point in the curves for the model
from \citet{steinberg} corresponds to $\gamma = 0$, that is, the regression
model where the constraint is inactive. No further reductions of the penalty
encoding the fairness are possible: since the model from \citet{steinberg} does
not outperform FRRM even then, we conclude that FRRM dominates it even for
larger values of $r$.

\citet{reductions} estimate a fair classifier by choosing an optimal model over
the set $\Delta$ of randomised classifiers subject to an inequality constraint
that enforces fairness:
\begin{align}
  &\min_{Q \in \Delta} \mathrm{err}(Q)&
  &\text{subject to}&
  &\mathbf{M} \boldsymbol{\mu}(Q) \le \mathbf{c} + \boldsymbol{\varepsilon}
\label{eq:agarawal}
\end{align}
where $\boldsymbol{\mu} \in \mathbb{R}^{|\mathcal{K}|\times|\mathcal{I}|}$ and
$\mathbf{c} \in \mathbb{R}^{|\mathcal{K}|}$ describe the linear constraints for
the chosen definition of fairness, $\mathbf{M}(Q) \in
\mathbb{R}^{|\mathcal{I}|}$ is a vector of conditional moments of functions of
the classifier $Q$ and $\boldsymbol{\epsilon}\in\mathbb{R}^{|\mathcal{K}|}$
controls the level of fairness. The choice of $\mathbf{M}, \boldsymbol{\mu}$ and
$\mathbf{c}$ is subject to the chosen definition of fairness. The solution to
\mref{eq:agarawal} is found through a series of cost-sensitive classification
problems by rewriting it as a saddle point problem with a Lagrangian multiplier
and applying the exponentiated gradient reduction proposed by \cite{freund} and
\cite{kivinen}.

\citet{reductions} provide an implementation of their approach in
\emph{Fairlearn} \citep{fairlearn}. It supports both regression and
classification but it does not support statistical parity for regression: hence
we compare it only with FGRRM. To match \emph{fairml}, we use as base classifier
the logistic regression of \emph{scikit-learn} with no penalty and ``newton-cg''
solver. For computational reasons, we only evaluate $\epsilon = 0.1$, which was
also considered in \citet{reductions}, and we fix all other parameters to their
default values. Furthermore, \emph{Fairlearn} only allows a single categorical
sensitive attribute: this is not a limitation for the COMPAS data (we merged the
two sensitive attributes into a single variable), but it is for the the ADULT
data (we used sex as the only sensitive attribute). Finally, \emph{Fairlearn}
did not converge for the BANK data. The average F1 we obtain for both data sets
($\mathrm{F1} = 0.6787$ for COMPAS and $\mathrm{F1} = 0.5805$ for ADULT) is
smaller than those produced by FGRRM ($[0.6891, 0.7029]$ for COMPAS and
$[0.6141, 0.6489]$ for ADULT) over all the considered values of $r$ .

\section{Conclusions}
\label{sec:conclusions}

In this paper we presented a general framework for learning fair regression
models that comprises both linear and generalised linear models. Our proposal,
which we call F(G)RRM for \emph{fair (generalised) ridge regression models},
uses a ridge penalty to reduce the proportion of variance (deviance) explained
by the sensitive attributes over the total variance (deviance) explained by the
model. Unlike most other approaches in the literature, it F(G)RRM can handle
arbitrary types and combinations of predictors and sensitive attributes and
different types of response variables.

Compared to the other approaches we have considered, we show that F(G)RRM
achieve a better predictive accuracy and a better goodness of fit for the same
level of fairness. (This is despite the optimality guarantees of NCLM, which we
show may not hold in practical applications.) In addition, we argue that F(G)RRM
produces regression coefficient estimates whose behaviour is more intuitive than
the other models we investigated in this paper.

F(G)RRM compares favourably with NLCM and ZL(R)M in two other respects as well.
Firstly, it is mathematically simpler and easier to implement since the only
numeric optimisation it requires is root finding in a single variable bounded in
a finite interval; the coefficient estimates are either available in closed form
(for FRRM) or can be estimated with standard software (for FGRRM). Secondly,
F(G)RRM is more modular than NLCM and ZL(R)M: it can be extended to use kernels
for modelling nonlinear relationships, different penalties, and different
definitions of fairness. It can accommodate multiple definitions of fairness
simultaneously as well.

\begin{acknowledgements}
Marco Scutari and Manuel Proissl acknowledge the UBS-IDSIA research
collaboration for the advancement of financial services with Artificial
Intelligence, which served as the host for this joint work on algorithmic
fairness. Francesca Panero was supported by the EPSRC and MRC Centre for
Doctoral Training in Statistical Science, University of Oxford (grant
EP/L016710/1).
\end{acknowledgements}

% \bibliographystyle{spbasic}
% \bibliography{bibliography}

\end{document}